\documentclass[runningheads]{llncs}

\usepackage[mobile]{eccv}

\usepackage{eccvabbrv}

\usepackage{graphicx}
\usepackage{booktabs}
\usepackage{multirow}
\usepackage{hhline}
\usepackage{makecell}
\usepackage{bm}
\usepackage{wrapfig}
\usepackage{bbding}
\usepackage{caption}
\usepackage{mathrsfs}
\usepackage{framed}
\usepackage{makecell}
\usepackage{color}
\usepackage{colortbl}  
\usepackage{xcolor}

\usepackage[ruled]{algorithm2e}

\usepackage[accsupp]{axessibility}  

\definecolor{Blue1}{RGB}{23, 178, 251}

\makeatletter
\renewcommand\@fnsymbol[1]{%
    \ifcase#1\or
    *\or 
    \textdagger\or
    \textdaggerdbl\or
    \fi
}
\makeatother

\usepackage{hyperref}

\usepackage{orcidlink}

\begin{document}

\title{Towards Reliable Advertising Image Generation Using Human Feedback} 


\author{Zhenbang Du\inst{1, 2}\thanks{Equal contribution}\thanks{Work done while interning at JD.com}\orcidlink{0000-0002-1386-8381} \and
Wei Feng\inst{2^*} \and Haohan Wang\inst{2} \and Yaoyu Li\inst{2}\and Jingsen Wang\inst{2} \and  Jian Li\inst{2} \and  Zheng Zhang\inst{2} \and Jingjing Lv\inst{2} \and Xin Zhu\inst{2} \and Junsheng Jin\inst{2} \and Junjie Shen\inst{2} \and Zhangang Lin\inst{2} \and Jingping Shao\inst{2}}

\authorrunning{Z.~Du et al.}

\institute{School of Artificial Intelligence and Automation,\\ Huazhong University of Science and Technology, Wuhan, China \\
\email{dzb99@hust.edu.cn} \and
 Retail Platform Operation and Marketing Center, JD, Beijing, China \\
 \email{\{fengwei25, wanghaohan1, liyaoyu1, wangjingsen, lijian21, zhangzheng11, lvjingjing1, zhuxin3, jinjunsheng1, shenjunjie, linzhangang, shaojingping\}@jd.com}}

\maketitle

\begin{abstract}
    In the e-commerce realm, compelling advertising images are pivotal for attracting customer attention. While generative models automate image generation, they often produce substandard images that may mislead customers and require significant labor costs to inspect. This paper delves into increasing the rate of available generated images. We first introduce a multi-modal Reliable Feedback Network (RFNet) to automatically inspect the generated images. Combining the RFNet into a recurrent process, Recurrent Generation, results in a higher number of available advertising images. To further enhance production efficiency, we fine-tune diffusion models with an innovative Consistent Condition regularization utilizing the feedback from RFNet (RFFT). This results in a remarkable increase in the available rate of generated images, reducing the number of attempts in Recurrent Generation, and providing a highly efficient production process without sacrificing visual appeal. We also construct a Reliable Feedback 1 Million (RF1M) dataset which comprises over one million generated advertising images annotated by human, which helps to train RFNet to accurately assess the availability of generated images and faithfully reflect the human feedback. Generally speaking, our approach offers a reliable solution for advertising image generation. Our dataset and code are available at \url{https://github.com/ZhenbangDu/Reliable_AD}. 
  \keywords{Diffusion Model \and Human Feedback \and E-commerce}
\end{abstract}

\section{Introduction}
\label{sec:intro}

An attractive advertising image is essential for e-commerce success, as it can lead to a higher click-through rate (CTR) \cite{wang2021ctr}.  Manual creation requires significant labor costs, therefore the demand for automatic advertising image production is on the rise.  However, previous approaches \cite{chen2021efficient, chen2021automated, mishra2020learning, wei2022towards} often result in visual mismatches between the product and background. Advanced diffusion models \cite{ho2020denoising} offer a remedy. A combination with ControlNet \cite{zhang2023controlnet} has shown promise in producing harmonious backgrounds for products while keeping the details of products identical for advertising use \cite{wang2023generate, Li2023PlanningAR}.

\begin{figure}[tbp]    \centering
  \includegraphics[width=0.95\columnwidth]{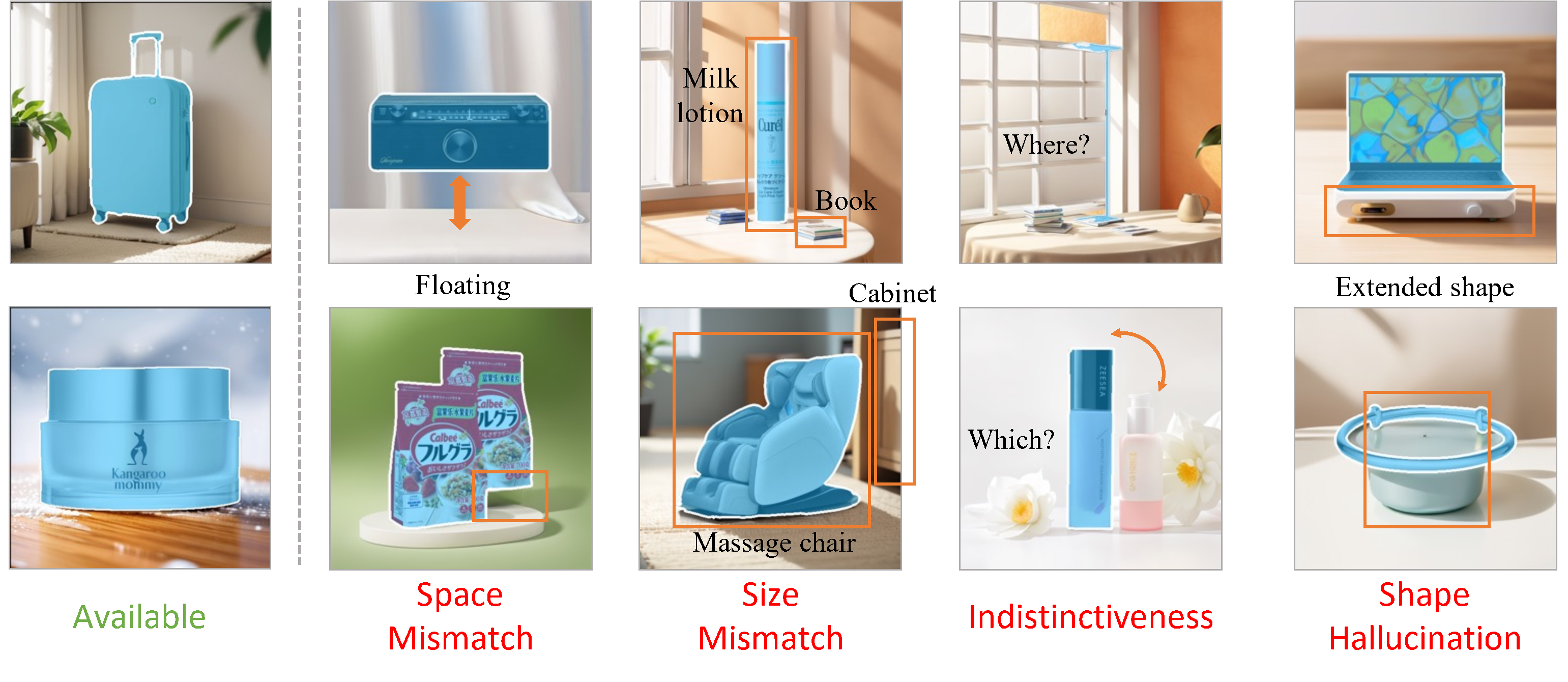}
  \caption{The available generated advertising images and different types of bad cases. The products are highlighted by \textcolor{Blue1}{blue} masks. Bad cases bring misleading information, \eg, the unrealistic sizes or shapes of products, and customers may have difficulty discerning the products in images.} 
  \label{Fig1}
\end{figure}

Despite the potential of generative models to create appealing backgrounds, we have observed a frequent production of substandard advertising images as depicted in \cref{Fig1}, which encompass various cases such as space and size mismatches, indistinctiveness, and shape hallucination. These flawed images can lead to customer misunderstandings about products and bring a subpar shopping experience, therefore considerable labor is necessitated to inspect the generated images. Such drawbacks limit the broader application of generative models in advertising image production. So the core problem is \textit{a low rate of available images}. It poses a new challenge to us how to establish a reliable advertising image generation pipeline capable of producing images with a high available rate. 

A natural solution is to generate images repeatedly until an available image is obtained (Recurrent Generation) due to the randomness in generation. To substitute labor inspection in this repetitive process, a novel \underline{R}eliable \underline{F}eedback \underline{Net}work (\textbf{RFNet}) is introduced to act as a human inspector, assessing the availability of generated advertising images. As simply depending on a single generated image, the model can not effectively obtain the knowledge pivotal to the precise inspection, \eg, what the product is and how the product appears in the background. So RFNet integrates multiple auxiliary modalities to provide critical information contributing to the judgment of different unavailable cases. Meanwhile, we construct a large-scale dataset called \underline{R}eliable \underline{F}eedback \underline{1} \underline{M}illion dataset (\textbf{RF1M}), which includes over one million elaborate generated advertising images with rich human annotations, playing a crucial role in training RFNet to mirror human feedback accurately.

While Recurrent Generation greatly increases the number of available images, multiple attempts will dramatically prolong the generation process owing to the inherent poor ability of generated models. Using human feedback to enhance the capability of diffusion models \cite{ouyang2022training, Touvron2023Llama2O} provides a viable option, which has successfully improved the visual quality of generated images \cite{Fan2023DPOKRL, Wallace2023DiffusionMA, Xu2023ImageRewardLA}. However, the visual quality and availability of the generated images show a trade-off relationship, \eg, products with repetitive and simple backgrounds gain a high available rate yet low aesthetics. To tackle this, we propose a novel loss term, Consistent Condition (CC) regularization $L_{CC}$ to counteract uniformity and degradation of the generated backgrounds, circumventing the adversarial nature of conventional Kullback-Leibler (KL) regularization \cite{Fan2023DPOKRL, Wallace2023DiffusionMA}. Utilizing this regularization term, the feedback on generated images' deviation from the available type assessed by RFNet is directly back-propagated to fine-tune the diffusion model (\textbf{RFFT}). Our approach significantly enhances the available rate of generated advertising images without altering their aesthetics, offering a comprehensive solution to the challenge of reliable advertising image generation.

Our main contributions include:
\begin{itemize}
    \item An advertising image generation solution Recurrent Generation alongside the novel multi-modal model, \textbf{RFNet}, which simulates human feedback and effectively utilizes various modalities to help attribute fine-grained issue types.
    \item A straightforward and effective approach, \textbf{RFFT}, to refine the diffusion model using human feedback, along with an innovative Consistent Condition regularization to prevent collapse.
     \item A large-scale multi-modal dataset, \textbf{RF1M}, comprising over one million generated advertising images with rich annotations.
\end{itemize}

\section{Related Work}

In this section, we review prior works on advertising image generation and the use of human feedback to refine diffusion models.

\subsection{Advertising Image Generation}

Automatic advertising image generation offers significant convenience for e-commerce. Current approaches generally fall into two folds, template-based and generation-based. Template-based approaches \cite{chen2021efficient, chen2021automated, mishra2020learning, wei2022towards} typically utilized predefined templates to assemble various elements into images. However, such approaches failed to achieve aesthetic harmony between the product and its background, resulting in a discernible disconnection. Additionally, designing a variety of templates was costly. To counter this, generation-based approaches \cite{ku2023staging, wang2023generate, Li2023PlanningAR} employed GANs \cite{goodfellow2014generative, isola2017image} or advanced text-to-image diffusion models \cite{ho2020denoising, song2020score, Rombach_2022_CVPR, song2020denoising} to produce backgrounds that coordinate with the appearance of products, \eg, Wang \etal \cite{wang2023generate} made use of the inpainting technique \cite{Lugmayr2022inpaint} with product masks to produce personalized backgrounds. Although elaborate and coordinated advertising images were obtained by these approaches, the bad cases that happened in the generation process could result in confusion and complaints of customers. We tackle this issue with a Recurrent Generation strategy, complemented by the accurate inspection capability of RFNet. Additionally, our RFFT enhances the reliability of the diffusion model and further promotes the efficiency of the production process.

\subsection{Refining Diffusion Models with Human Feedback}

Reinforcement Learning from Human Feedback (RLHF) employs human-derived feedback to fine-tune models, aiming for outcomes that human prefers \cite{casper2023open, christiano2017deep, losey2022physical, ibarz2018reward, Touvron2023Llama2O, ouyang2022training, Ziegler2019FineTuningLM}. In aligning diffusion models with human preferences, RLHF plays a crucial role \cite{fan2023optimizing, lee2023aligning, Lee2024ParrotPM, Zhang2024LargescaleRL, rafailov2023direct}. Works such as DDPO \cite{Black2023TrainingDM} and DPOK \cite{Fan2023DPOKRL} treated the denoising process as a multi-step Markov decision process. They used a Policy Gradient \cite{Schulman2017ProximalPO} approach to update model parameters based on feedback from pre-trained reward models. Similarly, D3PO \cite{Yang2023UsingHF} and Diffusion-DPO \cite{Wallace2023DiffusionMA} enhanced diffusion models using human comparison data, avoiding the need for reward model training. Contrary to the reinforcement learning strategy, ReFL \cite{Xu2023ImageRewardLA} and DRaFT \cite{Clark2023DirectlyFD} directly fine-tuned diffusion models using the gradient of differentiable rewards in an efficient end-to-end manner \cite{Wallace2023EndtoEndDL}. Our work pioneers in using human feedback to establish a reliable advertising image generation solution. Additionally, our approach addresses the collapse during fine-tuning, resulting in a high available rate of generated images without compromising their appearance.

\section{Dataset}
The Reliable Feedback 1 Million (\textbf{RF1M}) dataset, constructed through meticulous human feedback, serves as a pivotal resource for inspecting and improving the generation of advertising images. Compared to prevailing large-scale image generation dataset Laion-5B \cite{schuhmann2022laion}, DiffusionDB \cite{wang2022diffusiondb} and Pick-a-Pic \cite{kirstain2024pick}, RF1M is specifically designed for the advertising domain, addressing the acute need for expansive data resources in this field. Here’s an in-depth look at its composition, annotation, and potential impact on the community.

\begin{figure}[tbp]    \centering
  \includegraphics[width=0.95\columnwidth]{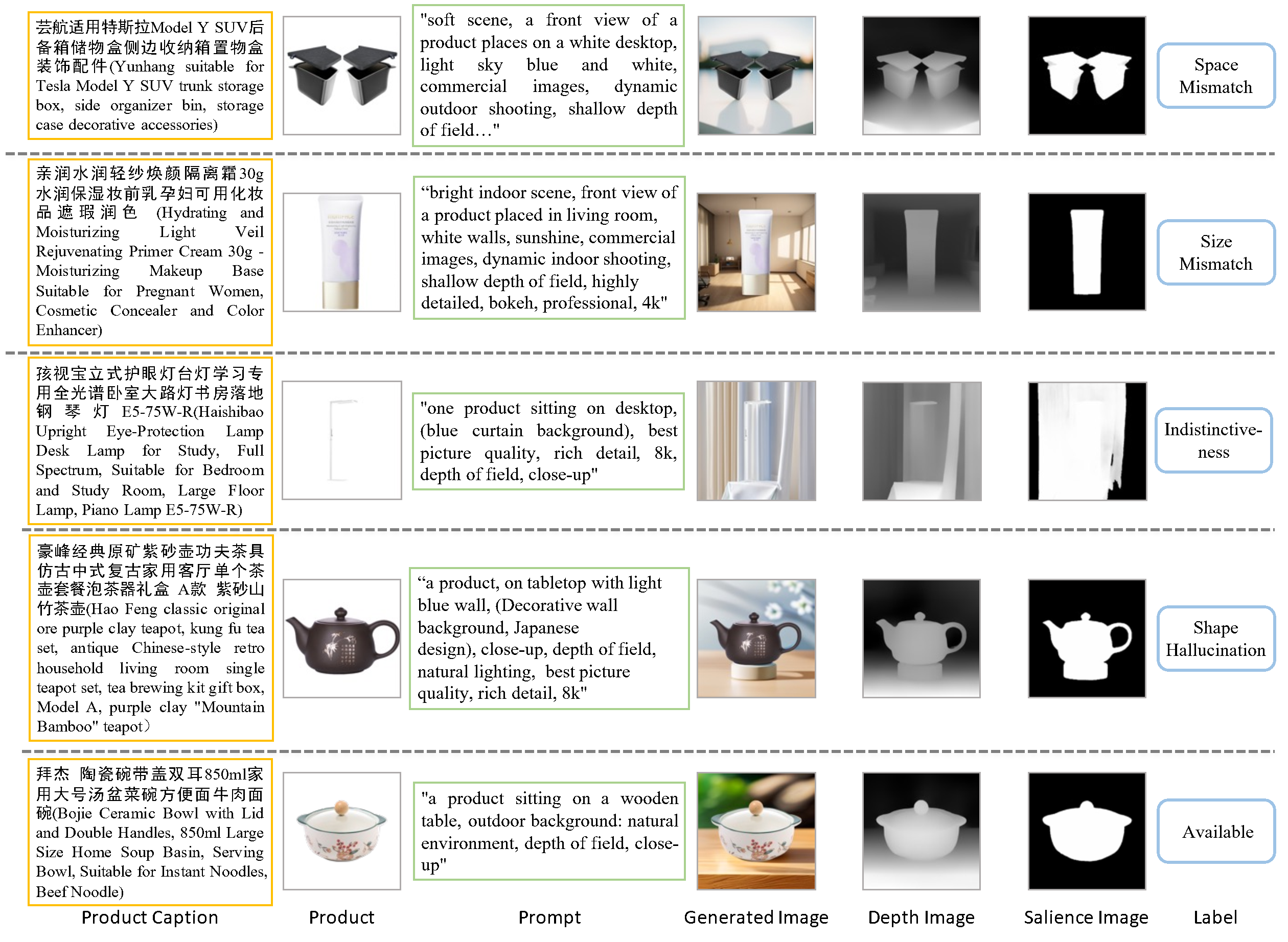}
  \caption{Some examples in RF1M. Each comprises rich annotations. The translations of Chinese captions are in the brackets.} 
  \label{fig:dataset}
\end{figure}

\noindent\textbf{Composition.} The dataset is generated using a collection of extensive products from JD.com. It encompasses 1,058,230 samples, each consisting of a variety of components aimed at providing a comprehensive understanding of advertising image generation:
\begin{itemize}
    \item The generated advertising image with corresponding transparent background product image, and carefully designed prompts by professional designers.
    \item Depth and salience images created by the dense prediction transformer \cite{ranftl2021vision} and U2-Net \cite{Qin_2020_PR} trained on e-commerce data, along with product caption, which assists in inspecting the availability of generated advertising image.
    \item Human-annotated label indicating the availability of image for advertising use.
\end{itemize}

These elements collectively offer a rich foundation for analyzing the generated advertising images. Some examples are exhibited in Fig. \ref{fig:dataset}.
 
\noindent\textbf{Annotation.} The annotators involved are well-versed in advertising and possess a deep understanding of the standards for advertising images. Within the dataset, samples have been further classified into five fine-grained categories as illustrated in Fig. \ref{Fig1}:
\begin{itemize}
    \item \textbf{Available.} Images deemed suitable for advertising purposes.
    \item \textbf{Space Mismatch.} Images where the product and background have inappropriate spatial relations, such as a part of the product is floating.
    \item \textbf{Size Mismatch.} Discrepancies between the product size and its background, \eg, a massage chair appears smaller than a cabinet.
    \item \textbf{Indistinctiveness.} Images where the product fails to stand out due to background complexity or color similarities.
    \item \textbf{Shape Hallucination.} Backgrounds that erroneously extend the product shape, adding elements like pedestals or legs.
\end{itemize}

\noindent\textbf{Potential Impact.} RF1M, with its multi-modal design and comprehensive features, is the cornerstone of the RFNet training and RFFT, and would further significantly impact the realm of e-commerce advertising and beyond, featuring three highlights: 

\begin{enumerate}
    \item \textbf{Large scale.} With its vast array of product categories and image types, RF1M surpasses previous advertising datasets BG60k \cite{wang2023generate}, PPG30k \cite{Li2023PlanningAR} and human feedback dataset ImageRewardDB \cite{Xu2023ImageRewardLA}, RichHF-18K \cite{liang2023rich}, while has comparable size to Pick-a-Pic \cite{kirstain2024pick}, providing a robust base for RFNet to mirror human feedback accurately across the diverse advertising image generation tasks. 
    
    \item \textbf{Scalability.} The multi-modal nature of RF1M provides sufficient information to RFNet to make judgments precisely. Beyond generating reliable advertising images, it also supports tasks such as advanced image understanding and image matting. This flexibility ensures the dataset meets the changing demands and can be applied in numerous areas within and beyond advertising.

    \item \textbf{Visual appeal.} The prompts and generation models are carefully designed for the products' characteristics, so the images have satisfying aesthetics and can attract customers' attention. We conducted a one-week online A/B test done in JD.com, resulting in a 2.2\% increase in CTR from over 60 million exposures, underscoring the high quality of these images, which accurately captured user preferences.
\end{enumerate}

\begin{figure}[tbp]    \centering
  \includegraphics[width=0.97\columnwidth]{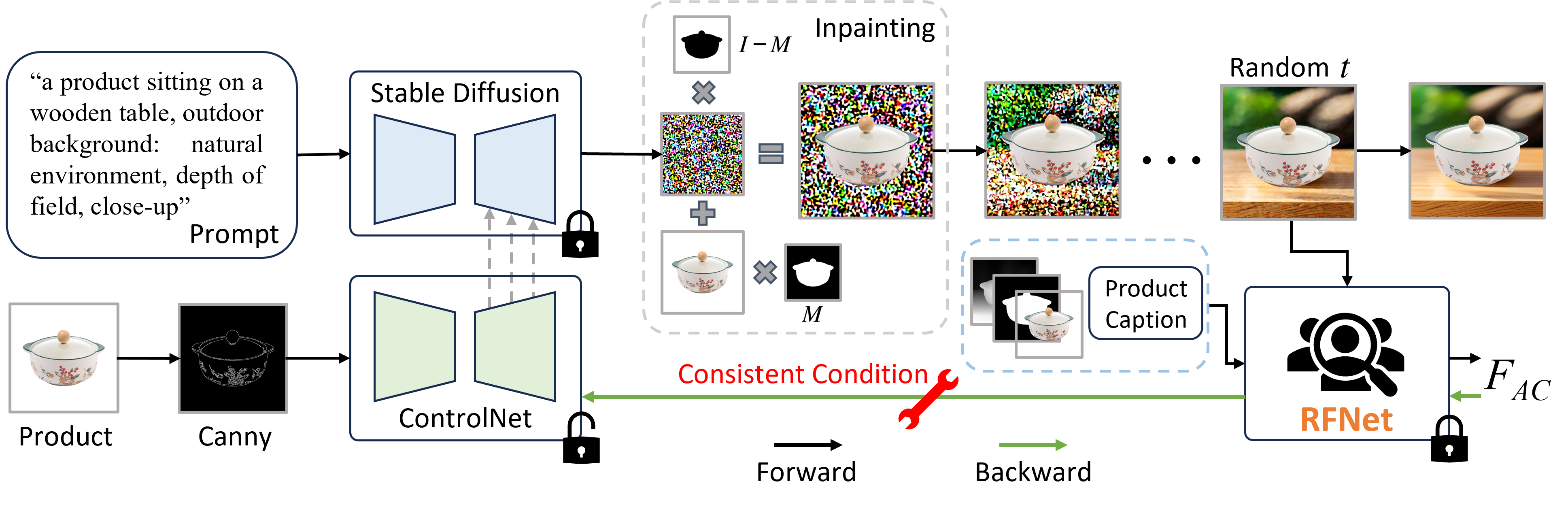}
  \caption{An overview of image generation-inspection pipeline. The advertising image is generated using product image and prompt by inpainting. And the feedback $F_{AC}$ provided by the RFNet is used to fine-tune the ControlNet with Consistent Condition regularization.} 
  \label{fig:Fig2}
\end{figure}

\section{Methodology}

\subsection{Preliminaries}
Our approach for generating advertising images is depicted in Fig. \ref{fig:Fig2}. We start with a text prompt describing the desired background and a product image $I_o$ with a transparent background. The prompt is input into Stable Diffusion \cite{Rombach_2022_CVPR}, and $I_o$ is pre-processed with canny edge detection before being fed into the ControlNet \cite{zhang2023controlnet}. We adopt DDIM \cite{song2020denoising} as our denoising schedule, the latent representation $x$ at step $t$ is calculated as following:
\begin{align}
x_{t}=\sqrt{\bar{\alpha}_{t}} \frac{x_{t+1}-\sqrt{1-\bar{\alpha}_{t+1}} \epsilon_\theta\left(x_{t+1}, t+1\right)}{\sqrt{\bar{\alpha}_t+1}}+\sqrt{1-\bar{\alpha}_{t}} \epsilon_\theta\left(x_{t+1}, t+1\right),
\label{eq1}
\end{align}
where $\epsilon_{\theta}$ represents the model \cite{Ronneberger2015unet, Rombach_2022_CVPR} that predicts added noise, and $\{\bar{\alpha}\}$ is a set of coefficients controlling the forward noise-adding process. To preserve the product's integrity and ensure a cohesive background, we employ an inpainting technique \cite{Lugmayr2022inpaint}. The latent representation $x_{t}$ is processed by:
\begin{align}
    x_{t}= (\bm{I}-M)\otimes x_{t} + M\otimes x_{o},
    \label{eq2}
\end{align}
where $x_{o}$ is the latent of $I_o$, $M$ is product mask, and $\otimes$ denotes the element-wise multiplication. After the $x_0$ is obtained, this latent is converted to the generated image $I_g$.

\subsection{Recurrent Generation with RFNet}
\label{sec:checker}
\begin{figure}[tbp]    \centering
  \includegraphics[width=0.97\columnwidth]{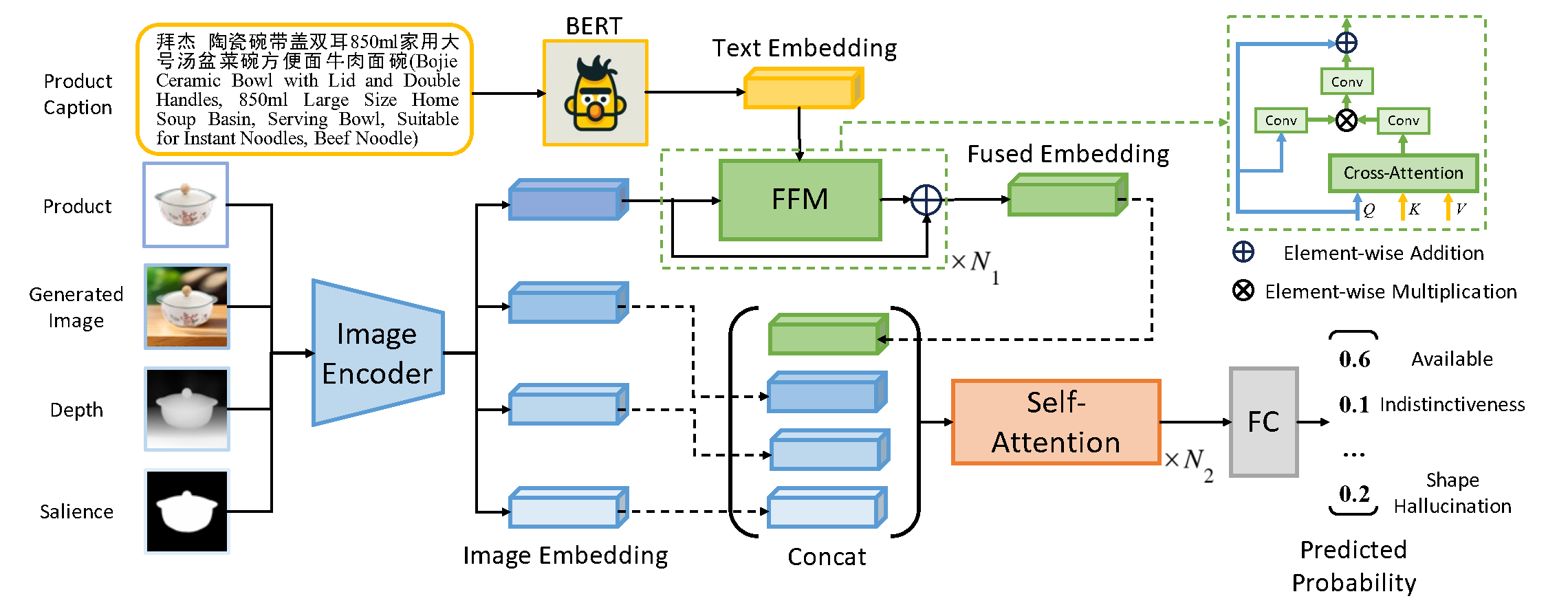}
  \caption{The proposed RFNet. Multiple auxiliary modalities contribute to the final inspection. The translation of the Chinese caption is in the brackets.} 
  \label{fig:checker}
\end{figure}

Due to the inherent randomness, repeated generation can significantly expand the number of available images. To automate the inspection process and eliminate human participation in this process, we introduce a multi-modal model, RFNet, to determine whether the generated image is available precisely, as illustrated in Fig. \ref{fig:checker}. In addition to $I_o$ and $I_g$, RFNet combines information from auxiliary modalities:
\begin{itemize}
    \item[$\bullet$] The depth image $I_d$ of $I_g$ produced by a depth estimation model, highlights the product's position relative to the background;
    \item[$\bullet$] The salience image $I_s$ of $I_g$, created by a salience detection model, outlines the product;
    \item[$\bullet$] The product caption $Cap$ which provides insight into the product's attributes.
\end{itemize}

$I_o$, $I_g$, $I_d$ and $I_s$ are fed into an image encoder to acquire respective image embeddings $\{\bm{e_o}, \bm{e_g}, \bm{e_d}, \bm{e_s}\}$. Concurrently, the $Cap$ is input into BERT \cite{Devlin2019bert} to obtain the text embedding $\bm{e_c}$,  aiding in recognizing the attributes of the product.

Since product captions often contain excessive information, \eg, brands, we focus on distilling the vision-related attributes from the caption. So $N_1$ Feature Filter Module (FFM)s are first employed, each consisting of a cross-attention layer and several convolution layers. The output of FFM is formulated as: 
\begin{align}
    \bm{e_f} = \text{Conv}\left(\text{Conv}(\text{CrossAttention}(\bm{e_o}, \bm{e_c})) \otimes \text{Conv}(\bm{e_o})\right)+ \bm{e_o},
\end{align}
where $\bm{e_o}$ serves as $Query$ with $\bm{e_c}$ acting as both $Key$ and $Value$ in cross-attention layer, $\text{Conv()}$ denotes convolution layer with $1\times1$ kernels, and $\otimes$ signifies element-wise multiplication. This process ensures that critical information from the caption is effectively integrated with the image embeddings, augmenting the model's understanding of the product.

With the fused embedding $\bm{e_f}$, different features are further integrated through $N_2$ self-attention layers \cite{vaswani2017attention},
\begin{align}
    \bm{f} = \text{SelfAttention}(\text{Concat}(\bm{e_f}, \bm{e_g}, \bm{e_d}, \bm{e_s})),
\end{align}
where $\text{Concat()}$ stands for concatenation. These stacked layers capture the critical features across the embeddings. Finally, a fully-connected classifier determines each case's probability of the generated image. 

Trained on large-scale RF1M, RFNet assesses the availability of generated advertising images accurately by considering a comprehensive set of visual and textual features, and providing nuanced feedback. This capability, combined with the Recurrent Generation strategy (illustrated in \cref{sec:RG}), significantly increases the number of available generated images for advertising use in an automatic manner.

\subsection{RFFT with Consistent Condition regularization}
\label{sec:rbl}
Although Recurrent Generation could produce more available images in total, the inherited poor ability of the generative model results in a prolonged and inefficient production process, posing a great challenge to the application. Our end-to-end generation-inspection pipeline allows feedback gradients from RFNet to directly fine-tune the diffusion model, enhancing its capability. Specifically, our proposed RFFT selects a random step $t$ among last 10 steps during the 40-step denoising process to generate the $\hat{x}_0^t=\frac{x_{t+1}-\sqrt{1-\bar{\alpha}_{t+1}} \epsilon_\theta\left(x_{t+1}, {t+1}\right)}{\sqrt{\bar{\alpha}_{t+1}}}$ \cite{song2020denoising, Xu2023ImageRewardLA, prabhudesai2023aligning}. The resulting $\hat{x}_0^t$ is post-processed to $\hat{I}_g^t$ then inspected by RFNet to determine its availability, with feedback calculated as follows:
\begin{align}
    F_{AC} = -\frac{1}{N}\sum_{i=1}^{N} \bm{y}_d\text{log}(\bm{\hat{o}_i}),
\end{align}
where $\bm{y}_d$ is the one-hot vector representing the desired ``Available'' category, the vector $\bm{\hat{o}_i}$ holds the probability for each generated image case and $N$ is the total number of samples. The gradient $\nabla F_{AC}$ is then back-propagated to steer the model towards producing images with a higher probability of being available.
\begin{figure}[tbp]    \centering
  \includegraphics[width=0.95\columnwidth]{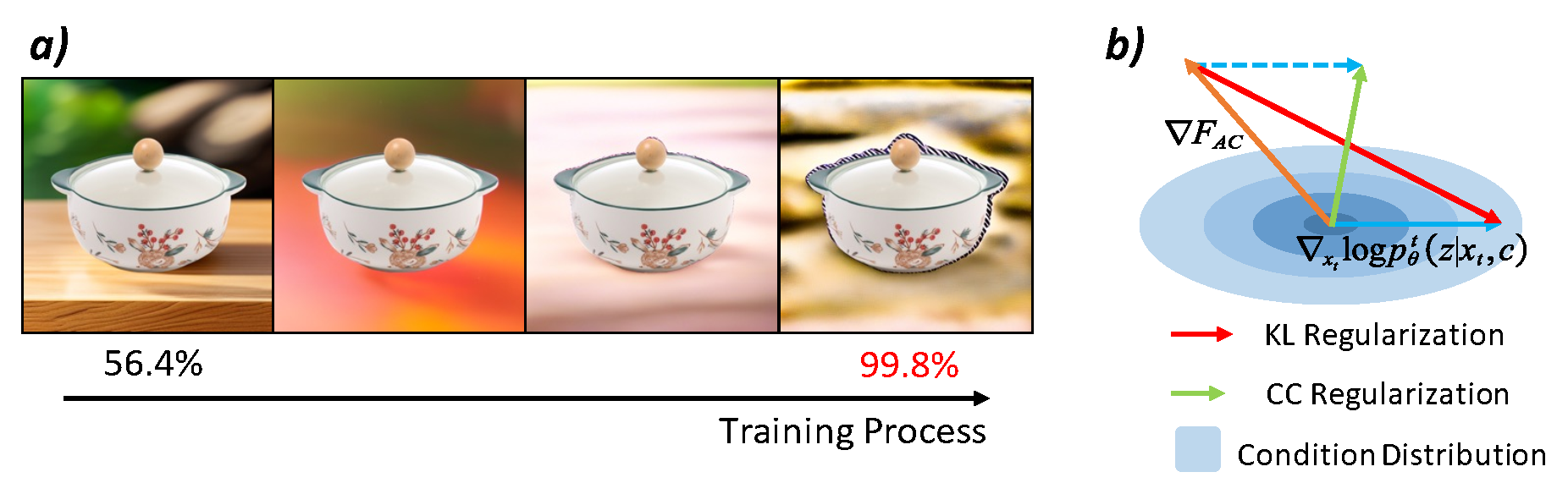}
  \caption{\textbf{a)} Background destruction happens in direct feedback backpropagation, where a high available rate (99.8\%) achievement with collapsed background. \textbf{b)} Comparison between KL regularization and our proposed CC regularization.} 
  \label{fig:hacking}
\end{figure}

Nevertheless, the training goal of making the model reliable runs counter to the aesthetics of the images generated by the model, \eg, repetitive and simple backgrounds can perfectly circumvent the bad cases we mentioned before. So as illustrated in Fig. \ref{fig:hacking}.a, as training progresses, the model achieves an extremely high available rate, yet yields homogeneous and aesthetically collapsed outputs. So a training approach that keeps the aesthetics of the image stable while increasing its availability rate is what we need. A prevalent solution involves the Kullback-Leibler (KL) regularization \cite{Fan2023DPOKRL, Wallace2023DiffusionMA}, a loss term that ensures the modified model does not diverge significantly from the desired distribution, thus maintaining diversity and preventing the convergence to sub-optimal, repetitive results, and this loss term can be formulated as:
\begin{align}
    L_{KL} = \text{KL}(p_{\theta}(\hat{x}_0^t|x_t,z,c)\Vert(p_{ref}(\hat{x}_0^t|x_t,z,c)),
\end{align}
where $c$ and $z$ are image and text control conditions, $p_{\theta}$ and $p_{ref}$ represent the distributions of current and reference models. However, as the feedback gradient endeavors to steer the image generation towards a higher available rate, the KL regularization strives to maintain the generated image unchangeable. This opposition mirrors the principles of adversarial training \cite{goodfellow2014generative, Ganin2015DomainAdversarialTO} and poses a challenge to achieve a win-win solution \cite{Ho2022ClassifierFreeDG}.

\begin{table}[htb]
\begin{minipage}[c]{0.5\textwidth}
  \caption{Inspection performance of different models.
  }
  \label{tab:tab1}
  \centering
  \resizebox{!}{0.25\columnwidth}{%
  \begin{tabular}{c|cccc}
    \Xhline{1px}
    Model & Precision & Recall & F1 & AP\\
    \hline
    ResNet50  & 74.87 & 73.66 & 74.26 &77.29\\
    ResNext50 & 77.73 & 76.88 & 77.30 &79.62 \\
    HRNet  & 72.89 & 73.12 & 73.01 &73.07\\
    ViT  & 75.59 & 78.33 &76.93 &79.31 \\
    \hline
    \cellcolor[HTML]{EFEFEF}{\textbf{Ours}}  & \cellcolor[HTML]{EFEFEF}{\textbf{86.45}} & \cellcolor[HTML]{EFEFEF}{\textbf{85.23}} & \cellcolor[HTML]{EFEFEF}{\textbf{85.83}} & \cellcolor[HTML]{EFEFEF}{\textbf{87.58}}\\
    \Xhline{1px}
  \end{tabular}}
\end{minipage}
\begin{minipage}[c]{0.5\textwidth}
\centering
\caption{Ablation study of RFNet.}
\resizebox{!}{0.28\columnwidth}{%
\begin{tabular}{@{}ccccc|c@{}}
\Xhline{1px}
$I_o$ & $I_g$ & $I_d$ & $I_s$ & $Cap$ & AP \\ \hline
      &  \Checkmark         &  \Checkmark     &  \Checkmark        &  \Checkmark       & 81.17   \\
\Checkmark      &           &  \Checkmark     &   \Checkmark       &  \Checkmark       & 82.06   \\
 \Checkmark   &   \Checkmark         &       &    \Checkmark       &   \Checkmark       &  85.31  \\
  \Checkmark     &    \Checkmark        &   \Checkmark     &          &   \Checkmark       & 83.91   \\
  \Checkmark  &  \Checkmark         &  \Checkmark     &   \Checkmark       &         & 84.53   \\
\Checkmark      &  \Checkmark         & \Checkmark       &  \Checkmark        &  \Checkmark       & \textbf{87.58}   \\ \hline
\multicolumn{5}{c|}{Coarse-grained} & 82.06  \\
\Xhline{1px}
\end{tabular}}
\label{tab:aba}
\end{minipage}
\end{table}

Instead of focusing on unchanged images, we aim to maintain visual quality. For text-to-image generation, the visual output is closely linked to the input text condition $z$ \cite{Witteveen2022InvestigatingPE}. In a classifier-free manner, we derive text guidance from the model's implicit classifier \cite{dhariwal2021diffusion, Ho2022ClassifierFreeDG, Nichol2021GLIDETP} by
\begin{align}
    \nabla_{x_t}\text{log}p_{\theta}^t(z|x_t, c) \approx - \frac{1}{\sqrt{1-\bar{\alpha}_{t}}}\Big(\epsilon_{\theta}(x_t, z, c) - \epsilon_{\theta}(x_t, c)\Big),
\end{align}
which indicates the direction where the text condition influences image generation. To ensure improvements in image availability do not compromise the core conditions, we introduce a Consistent Condition (CC) regularization term $L_{CC}$, as follows:
\begin{align}
    L_{CC} = \Vert \nabla_{x_t}\text{log}p_{\theta}^t(z|x_t, c) - \nabla_{x_t}\text{log}p_{ref}^t(z|x_t, c)\Vert_{2}.
\end{align}

Fig. \ref{fig:hacking}.b illustrates the advantage of $L_{CC}$ over $L_{KL}$. While $L_{KL}$ acts to limit updates from $\nabla F_{AC}$ potentially leading to rigidity, $L_{CC}$ offers a win-win approach. It maintains the direction of the condition, allowing for the model to be fine-tuned towards generating more available images. Thus the final feedback to fine-tune the diffusion model in RFFT is:
\begin{align}
    F_{total} = F_{AC}+\beta L_{CC},
\end{align}
where $\beta$ is a hyper-parameter.

\section{Experiments}

\subsection{Implementation Details}
For RFNet, we employ a ResNet50 \cite{He2015DeepRL}, pre-trained on ImageNet \cite{Deng2009ImageNetAL}, as the image encoder. RoBERTa \cite{Devlin2019bert, Liu2019RoBERTaAR}, fine-tuned on Chinese product descriptions, extracts text embeddings from product captions. We resize all images to $384\times384$ before encoding. The FFM and Self-Attention contain blocks of width 384 with 8 attention heads, and we set $N_1$ and $N_2$ to 1 and 3, respectively. The training spans 10 epochs, starting with a learning rate of 1e-4, which is reduced by a factor of 10 at epoch 5.

For diffusion model fine-tuning, we utilize 8 NVIDIA A100 GPUs, a local batch size of 4, and 4 gradient accumulation steps. We opt for AdamW \cite{Loshchilov2019adamw} with a learning rate of 1e-5. Unless specified otherwise, MajicmixRealistic\_v7 (Maji\_v7)\footnote{\url{https://civitai.com/models/43331/majicmix-realistic}} serves as the diffusion model, complemented by ControlNet V1.1 \cite{zhang2023controlnet}\footnote{\url{https://github.com/lllyasviel/ControlNet}}. During fine-tuning, only ControlNet is trained, while we freeze the remaining parameters. A 40-step DDIM with the last 10 steps chosen for fine-tuning is used \cite{Xu2023ImageRewardLA}.

\subsection{Advertising Image Inspection Performance}
\noindent\textbf{SOTA Approaches.} To validate the superiority of our proposed RFNet, we implement prevailing models ResNet50 \cite{He2015DeepRL}, ResNeXt50 \cite{Xie2016AggregatedRT}, HRNet \cite{Wang2019DeepHR} and Vision Transformer (ViT) \cite{Dosovitskiy2020AnII} to inspect the generated images.

\noindent\textbf{Evaluation Metrics.} Precision, recall, F1 score, and average precision (AP) are used to evaluate the prediction results of different models. We conduct the test on 1,000 images. It is noteworthy that since we are only concerned about the availability of the generated images in Recurrent Generation, we focus on whether the model is able to accurately identify the available images.

\noindent\textbf{Results.} Table \ref{tab:tab1} shows that RFNet outperforms across all metrics, highlighting the benefits of integrating multi-modal information and its effective structure. We conducted extensive experiments to evaluate the impact of various components within RFNet, presented in Table \ref{tab:aba}. Our experiments demonstrate the significant impact of each component within RFNet on the final AP, especially the crucial role of product images, whose ablation leads to a notable 6.41 drop in AP. We further trained the model using coarse-grained labels (Available/Unavailable), the decrease showcased that a fine-grained label helped to attribute the issues precisely, leading to improved performance.

\begin{figure}[tb]
\begin{minipage}[c]{0.55\textwidth}
  \centering
  \captionof{table}{Availability evaluation ($\%$) of different approaches using one attempt RG.}
  \resizebox{0.85\columnwidth}{!}{
  \begin{tabular}{c|cc}
    \Xhline{1px}
    Approach & Ava $(\uparrow)$  & Human Ava $(\uparrow)$ \\
    \hline
    Ori  & 56.4 & 70.1 \\

    PromptEng & 62.9 & 73.2 \\
    PPO  & 65.9 & 74.9  \\
    DPO  & 57.3 & 71.8  \\
    ReFL  & 84.7 & 84.9  \\
    \hline
     \cellcolor[HTML]{EFEFEF} \textbf{Ours}  &  \cellcolor[HTML]{EFEFEF} \textbf{85.5} &  \cellcolor[HTML]{EFEFEF} \textbf{86.3} \\

    \Xhline{1px}
  \end{tabular}}
  \label{tab:main_result}
\end{minipage}
\begin{minipage}[c]{0.45\textwidth}   
\centering
  \includegraphics[width=0.9\textwidth]{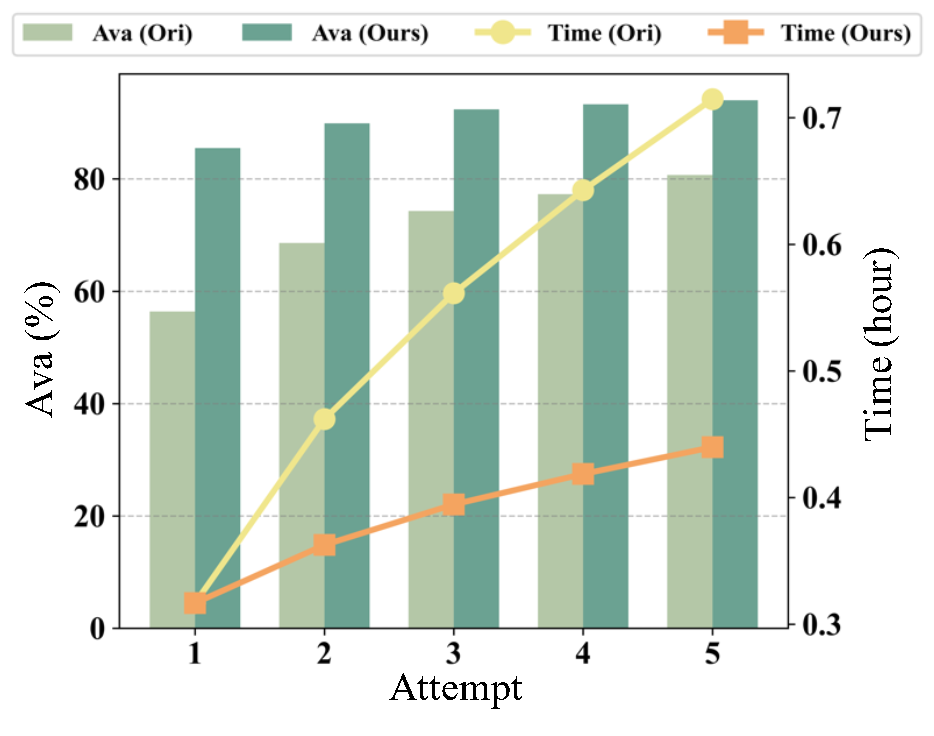}
  \captionof{figure}{Available rate ($\%$) and time (hour) for Ori and Ours with different attempts RG.} 
   
  \label{fig:rg}
\end{minipage}
\end{figure}

\begin{figure}[ht]
    \centering
    \includegraphics[width=0.9\textwidth]{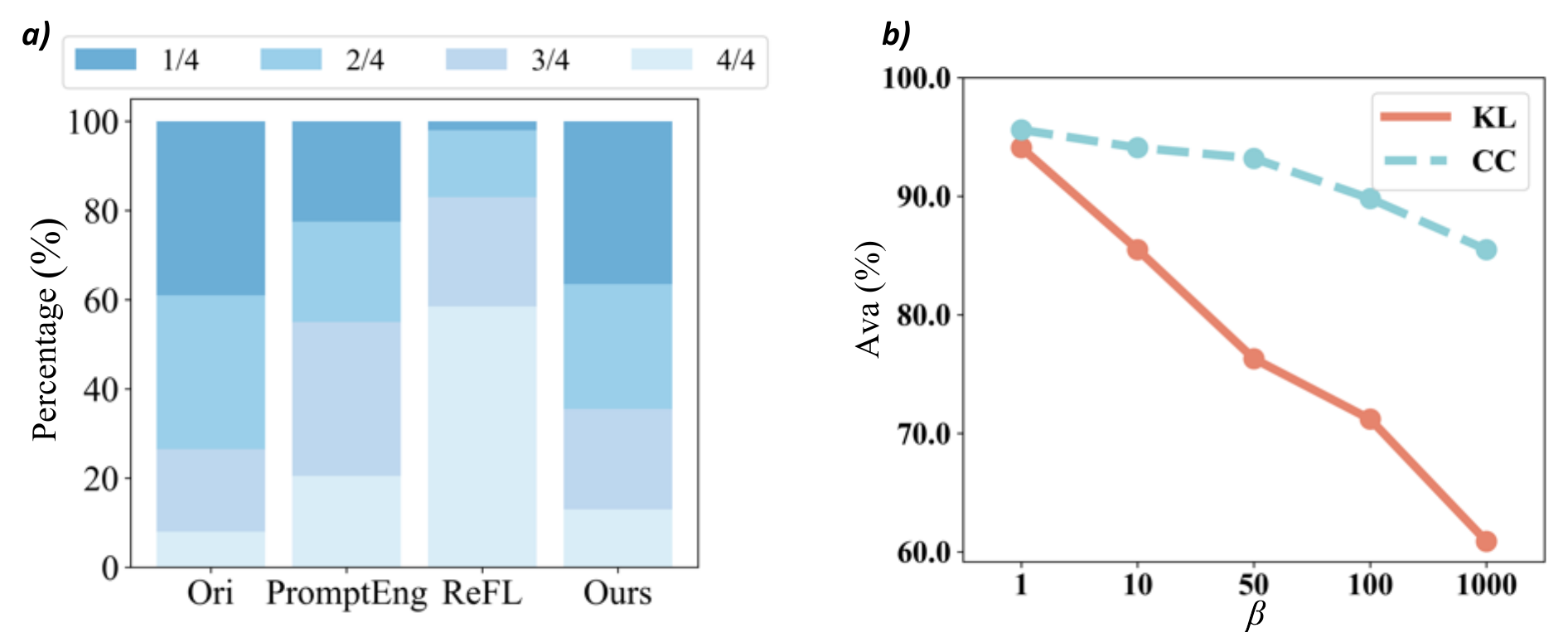} 
    \caption{\textbf{a)} Human rankings of different approaches; \textbf{b)} Available rate ($\%$) with different $\beta$ for $L_{KL}$ (KL) and $L_{CC}$ (CC).}
    \label{fig:plot}
\end{figure}

\subsection{Advertising Image Reliability Performance}
\label{sec:reli}

\noindent\textbf{SOTA Approaches.} 
To evaluate the effectiveness of refining diffusion models, our approach is compared with Ori (using the original model), Prompt Engineering (PromptEng) (using modified prompts)\footnote{Add ``irregular shape, extended shape, floating, table legs, pedestal, improper position, improper size, indistinct background'' to the negative prompt \cite{Armandpour2023ReimagineTN}.}, and SOTA RLHF approaches PPO \cite{Black2023TrainingDM, Fan2023DPOKRL, Lee2024ParrotPM, Zhang2024LargescaleRL}, DPO \cite{rafailov2023direct, Yang2023UsingHF, Wallace2023DiffusionMA}, and ReFL \cite{Xu2023ImageRewardLA, Clark2023DirectlyFD}.

\noindent\textbf{Evaluation Metrics.} 
For availability evaluation, we conduct the test on 1,000 products. The available rate is defined as the ratio of images deemed available by our RFNet to the total number of images inspected, and denoted by ``Ava''. To counter potential biases introduced by the RFNet, we also performed a human inspection which involved experienced advertising annotators assessing the available rate, denoted as ``Human Ava''. We calculate the total consuming time (hour) using an NVIDIA H800 GPU. For aesthetic assessment, we randomly select 200 available images with identical products, prompts, and generation seed for different approaches. 50 experienced advertising practitioners are employed to rank the images based on their preferences. We count the percentage of different rankings for distinct approaches. 

\noindent\textbf{Results.}
The outcomes highlight our approach's superior performance. Several conclusions could be made:

\begin{enumerate}
    \item From the result in Table \ref{tab:main_result}, our RFFT could get an extremely high available rate against other approaches. The same trend of ``Ava'' and ``Human Ava'' further demonstrates that RFNet reflects human feedback faithfully.
    \item As shown in \cref{fig:rg}, RG could greatly increase the rate of available images with multiple attempts. Thanks to the inherited strong ability of our model, it needs a shorter production time, shedding light on our RFFL to provide a reliable and efficient solution. 
    \item The preference assessment in \cref{fig:plot}.a demonstrates that RFFT could achieve a relative aesthetic quality compared to the original model, and gain a lot over ReFL owing to the utilization of CC regularization.
\end{enumerate}    

\begin{figure}[tbp]    \centering
  \includegraphics[width=0.85\columnwidth]{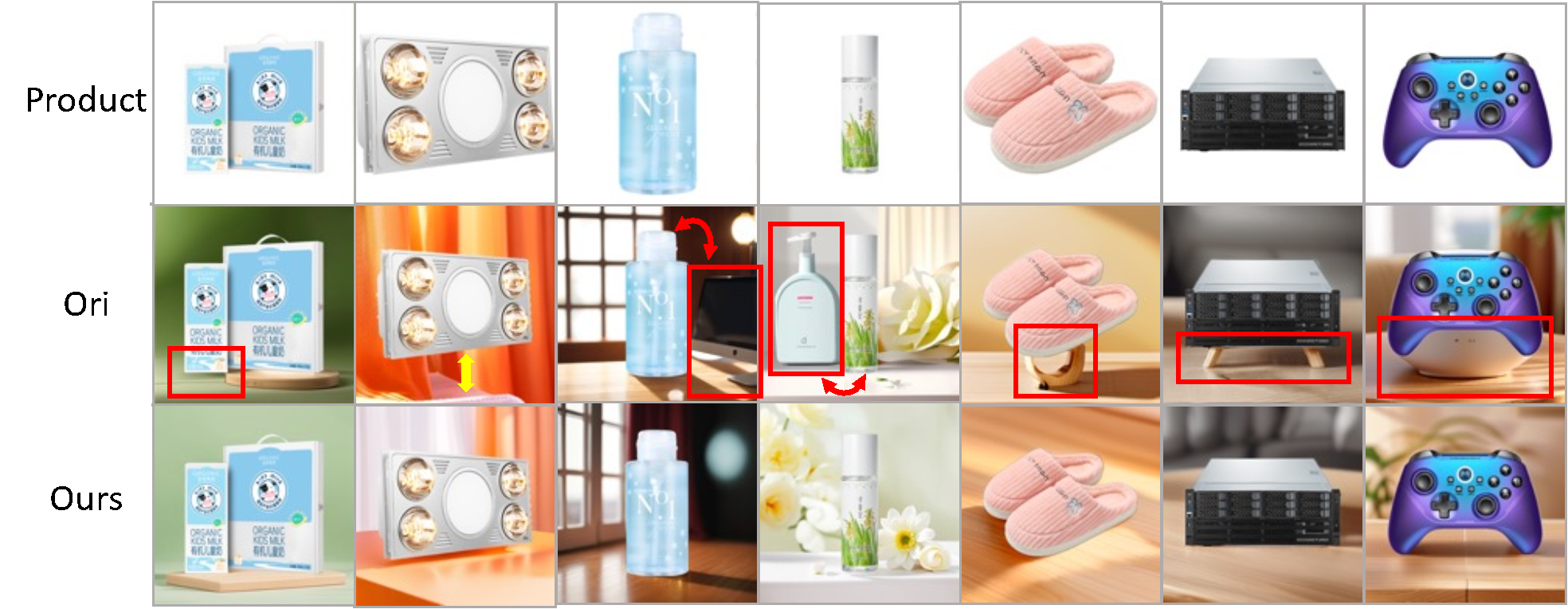}
  \caption{Comparison of generated advertising images by Ori and Ours.} 
  \label{fig:Fig7}
\end{figure}

These results underscore our approach's enhanced capability to increase the available rate and the producing efficiency while keeping the visual performance stable. Some bad cases solved by our approach are shown in Fig. \ref{fig:Fig7}.

\subsection{Comparison with KL Regularization}
We explore the impact of the hyper-parameter $\beta$ on both KL regularization $L_{KL}$ and our proposed CC regularization $L_{CC}$. The results, illustrated in Fig. \ref{fig:plot}.b, reveal a notable trend: as $\beta$ increases, the available rate significantly decreases under $L_{KL}$, highlighting its adversarial nature. Conversely, $L_{CC}$ demonstrates resilience to increased $\beta$ values, maintaining a higher available rate. This contrast underscores the effectiveness of $L_{CC}$ in circumventing adversarial effects.

\subsection{Generalization Performance}
\label{sec:gene}
To assess the flexibility of our RFFT, we examine the generalized capability of the fine-tuned ControlNet when integrated with various LoRAs \cite{2022Edwardlora} and diffusion model weights. This is critical as retraining the network for each new LoRA or diffusion model weights combination would be impractical. As depicted in Table \ref{tab:genera}, experiments demonstrate that the ControlNet, once refined, significantly enhances the available rate across different LoRAs and diffusion model weights, including Maji\_v6\footnote{\url{https://civitai.com/models/43331/majicmix-realistic}} and SD\_v1.5\footnote{\url{https://huggingface.co/runwayml/stable-diffusion-v1-5}}. This generalized ability reduces the need for repetitive training, and underscores broader application of RFFT.

\begin{figure}[ht]
    \centering
    \includegraphics[width=0.7\textwidth]{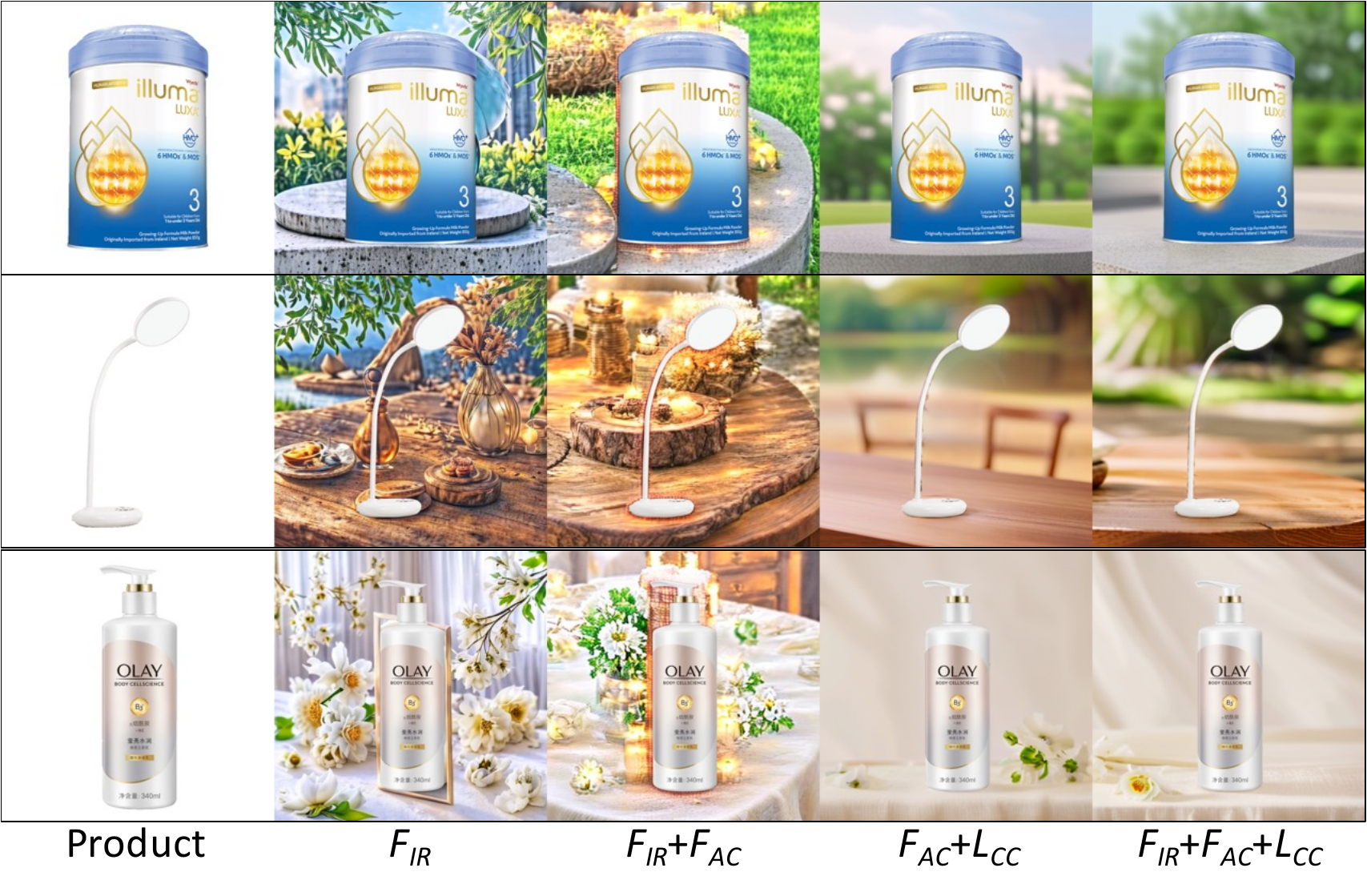} 
    \caption{Comparison of different feedback combination strategies.}
    \label{fig:combineIR}
\end{figure}

\subsection{Integration with Other Feedback}

We further incorporate ImageReward \cite{Xu2023ImageRewardLA} whose original goal is to improve aesthetics as an additional feedback, $F_{IR}$, during the RFFT process. The outcomes, both quantitative and qualitative, are presented in Table \ref{tab:combineIR} and Fig. \ref{fig:combineIR}, respectively. The following observations were made:

\begin{itemize}
\item The inclusion of $F_{IR}$, and the combination of $F_{IR}$ and $F_{AC}$, significantly enhance aesthetic appeal. However, these often result in excessive detail in the backgrounds, leading to obscured product features, which are undesirable for advertising purposes. Moreover, the combination of $F_{IR}$ and $F_{AC}$ results in collapsed textures despite achieving a high available rate.

\item Both $F_{AC}+L_{CC}$ and $F_{IR}+F_{AC}+L_{CC}$ configurations achieve high available rates. However, comparing these two, $F_{IR}$ does not significantly affect background aesthetics. This suggests that while $F_{IR}$ attempts to modify the background condition, $L_{CC}$ aims to maintain it. This side-by-side corroborates that our proposed $L_{CC}$ keeps the background visual quality unchanged, which is particularly important for our RFFL process.
\end{itemize}

\begin{figure}[htbp]
  
    \begin{minipage}[b]{0.6\textwidth}
    \centering
    \captionof{table}{Available rate of different diffusion model weights/LoRA. }
    \resizebox{\columnwidth}{!}{%
    \begin{tabular}{c|cc|c|c}
      \Xhline{1px}
    \multirow{2}{*}{Approach} & \multicolumn{2}{c|}{Maji\_v7} & \multirow{2}{*}{Maji\_v6} & \multirow{2}{*}{SD\_v1.5}\\ \hhline{|~|-|-|~|~|} 
                            & $\text{LoRA}_1$     & $\text{LoRA}_2$  & &             \\ \hhline{|-|-|-|-|-|}
                            Ori                      & 56.4             &   55.0            &    65.2           &   68.3            \\
                           
                            Ours &  \textbf{85.5}            &  \textbf{79.7}             &     \textbf{84.1}         & \textbf{84.0}              \\ 
                            \Xhline{1px}
    \end{tabular}
    }
    
    \label{tab:genera}
  \end{minipage}
  \begin{minipage}[b]{0.4\textwidth}
    \centering
    \captionof{table}{Available rate of different combination strategies with $F_{IR}$.}
    \resizebox{0.5\columnwidth}{!}{%
    \begin{tabular}{@{}ccc|c@{}}
\Xhline{1px}
$F_{IR}$ & $F_{AC}$ & $L_{CC}$ &Ava \\ \hline
\Checkmark   & & & 31.5   \\
\Checkmark    & \Checkmark  &  & 87.6 \\

 & \Checkmark  & \Checkmark &  85.5\\
 \Checkmark & \Checkmark  & \Checkmark & 81.4 \\
    \Xhline{1px}
\end{tabular}

    }
    
    \label{tab:combineIR}
  \end{minipage}
\end{figure}

\section{Conclusion}

In conclusion, our study addresses the critical need for reliable advertising image generation in e-commerce, where visual appeal directly influences customer engagement and sales. We introduce Recurrent Generation with the multi-modal RFNet, which faithfully reflects human feedback from our extensive RF1M dataset, establishing a foundation for creating a substantial number of available images for advertising. Our RFFT significantly enhances the generated images available rate, thereby boosting production efficiency. Moreover, our innovative Consistent Condition regularization strikes a balance between a high available rate and aesthetic quality. Ultimately, our work automates creative processes, reduces costs, and improves the customer experience, underscoring the potential of AI-driven tools to revolutionize e-commerce.

\bibliographystyle{splncs04}
\bibliography{egbib}

\newpage
\section*{Supplemental Material}

In this supplemental material, we first describe the construction of the RF1M dataset (\cref{sec:dataset}). The Recurrent Generation is described in \cref{sec:RG}. We then detail the human-involved experiments (\cref{sec:huamn}) and address our responsibilities towards human subjects (\cref{sec:respon}). The detailed configuration for image generation is outlined in \cref{sec:config}, followed by additional visualization results in \cref{sec:visual}. Finally, we discuss our ethical concerns (\cref{sec:ethi}) and future work (\cref{sec:future}).

\section{Dataset Construction}
\label{sec:dataset}
\subsection{Annotation Guidance}
\label{sec:anno}
\begin{figure*}[ht]    \centering
  \includegraphics[width=1.0\columnwidth]{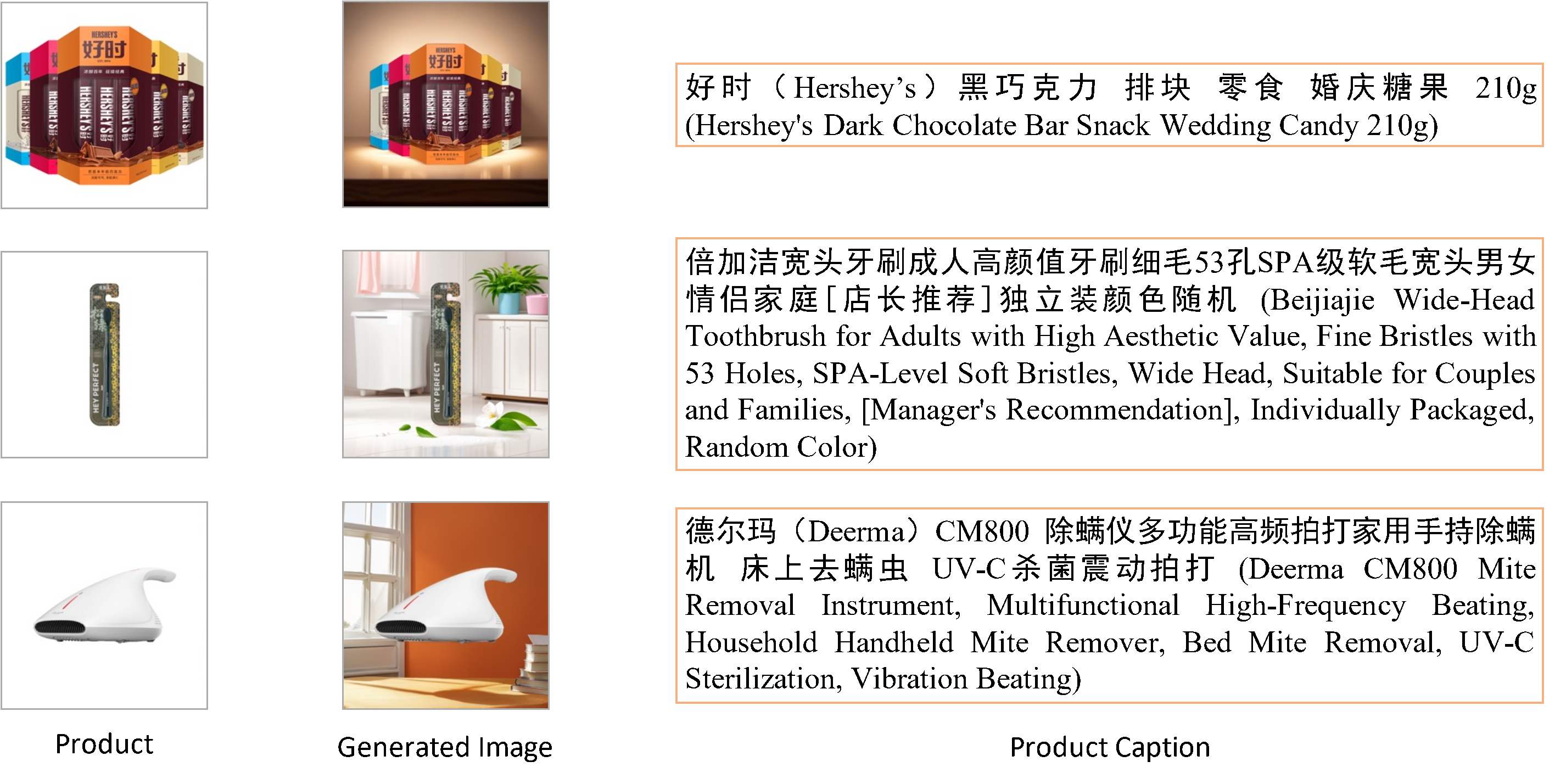}
  \caption{Some examples shown to annotators. The translations of Chinese captions are in the brackets.}
  \label{fig:anno}
\end{figure*}

During the \textbf{RF1M} annotation process, annotators are provided with the original product images, product captions, and generated images along with the following instructions:
\begin{framed}
Based on the product image and caption, determine the category of the advertising image:
\begin{enumerate}

\item \textbf{Space Mismatch.} Image where the product and background have inappropriate spatial relations contrary to physical laws, such as a part of the product is floating.
\item \textbf{Size Mismatch.} Discrepancies between the product size and its background violate common sense, \eg, a massage chair appears smaller than a cabinet.
\item \textbf{Indistinctiveness.} Image where the product fails to stand out due to background complexity or color similarities, and you cannot clearly figure out which or where the product is.  Additionally, this confusion might lead to misconceptions about the product itself, mistakenly suggesting the inclusion of extra freebies.
\item \textbf{Shape Hallucination.} Background that erroneously extends the product shape, adding elements like pedestals or legs, and giving you incorrect perceptions about the product.
\item \textbf{Available.} Image deemed available for advertising purposes, not falling into any of the above categories.
\end{enumerate}
\textbf{*} If the image belongs to several types simultaneously, \eg, the product has an inappropriate size and extended shape, annotate it with the most significant issue.
\end{framed}
And some examples to be annotated are shown in Fig. \ref{fig:anno}

\subsection{More Examples in Dataset}
\begin{figure*}[ht]    \centering
  \includegraphics[width=1.0\columnwidth]{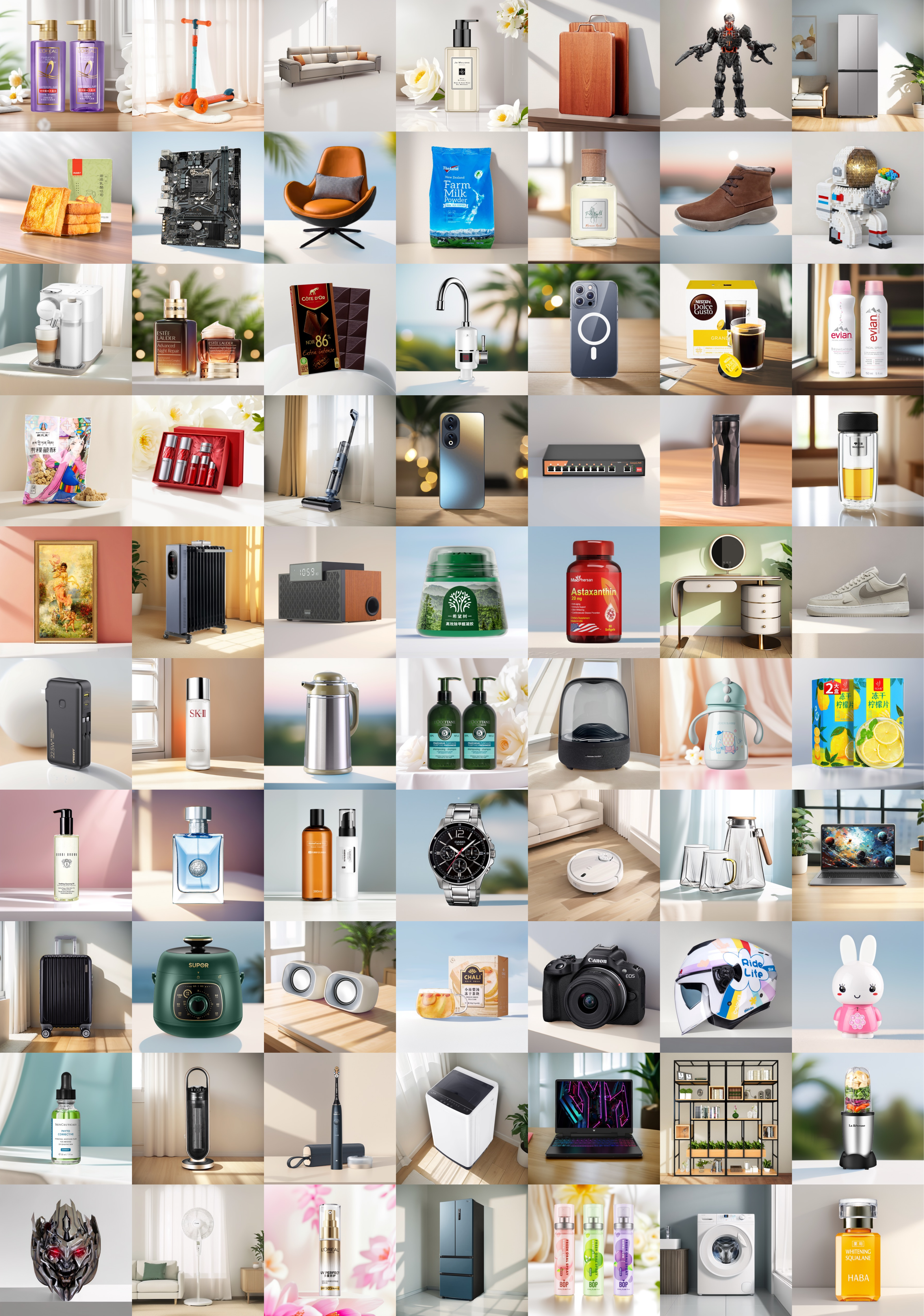}
  \caption{Some available generated images in the dataset, encompass various products in multiple scenarios.}
  \label{fig:avai}
\end{figure*}

\begin{figure*}[ht]    \centering
  \includegraphics[width=1.0\columnwidth]{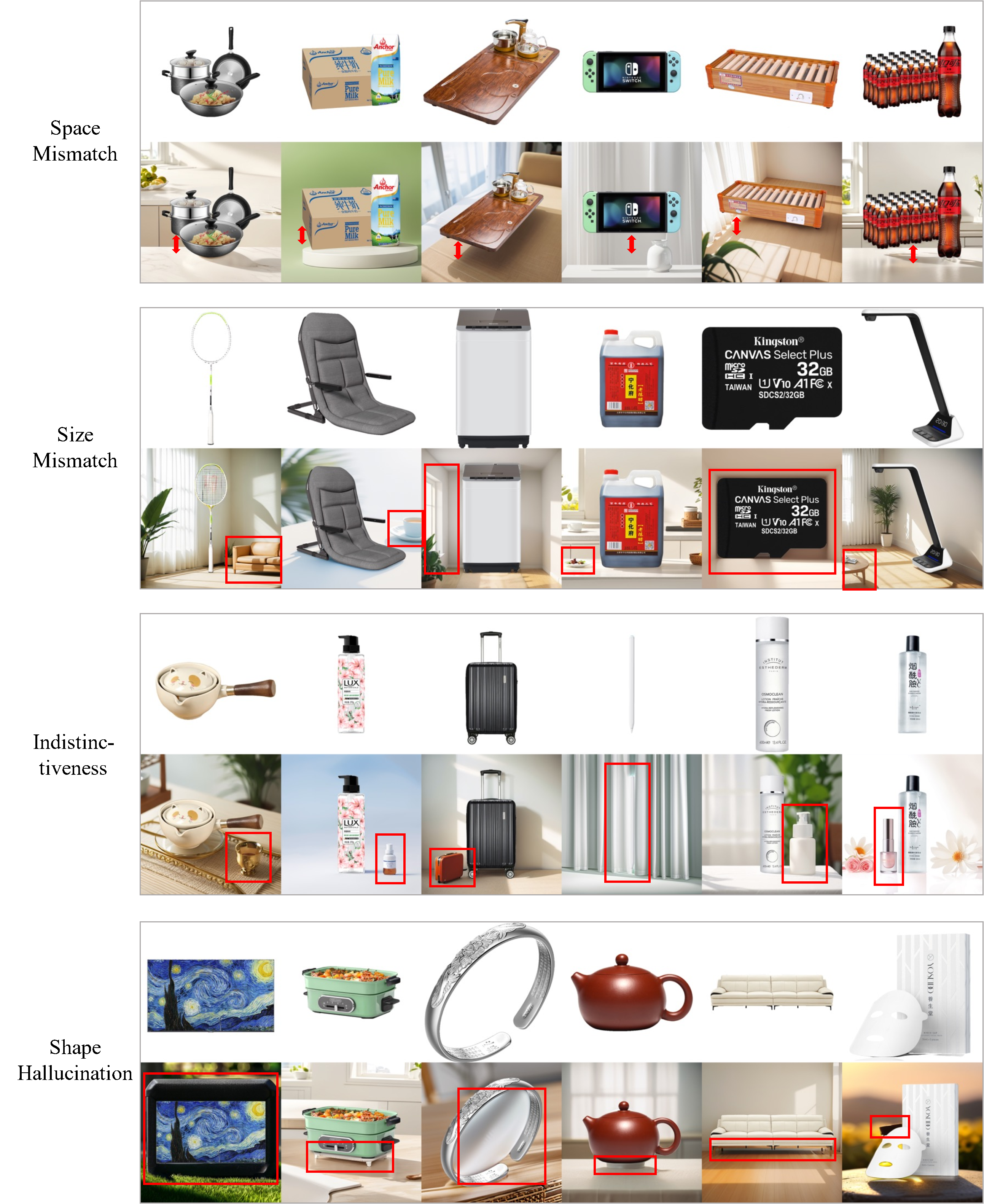}
  \caption{Some unavailable generated images in the dataset, encompass four bad cases.}
  \label{fig:unavai}
\end{figure*}

All samples in the RF1M dataset are generated using the approach outlined in paper, created with the original diffusion model before fine-tuning. We showcase additional examples of both available and unavailable generated images in Figs. \ref{fig:avai} and \ref{fig:unavai}. For brevity, auxiliary modalities such as product captions, depth images, salience images, and prompts are not shown. The dataset includes a broad variety of products and background scenes with a strong visual appeal.

\section{Recurrent Generation}
\label{sec:RG}
We show our Recurrent Generation strategy in \cref{al:1}.

\begin{algorithm}[t]
  \caption{Recurrent Generation}
  \label{al:1}
  \SetKwData{Left}{left}\SetKwData{This}{this}\SetKwData{Up}{up}
  \SetKwFunction{Union}{Union}\SetKwFunction{FindCompress}{FindCompress}
  \SetKwInOut{Input}{Input}\SetKwInOut{Output}{Output}
  \small
  \Input{Depth estimation model $Depth()$, Salience detection model $Salience()$, $RFNet()$, product image $I_o$ and caption $Cap$, denoising process $Denoise()$, max attempts $K$}

  attempt $k \leftarrow 0$, initial noise $x_{T}\sim\mathscr{N}(0, \bm{I})$;\\
  \While {$k < K$} {

  $x_0 = Denoise(x_{T})$; $I_g \leftarrow x_0$;  $\#$ latent to image

  $y_{pre} = RFNet(I_o, I_g, Depth(I_g), Salience(I_g), Cap)$;

  \eIf{$y_{pre}==\text{``Available''}$}{
   $I_{ava} = I_g$;

Break;}{$I_{ava} = None$;}

  $k \leftarrow k + 1$;
  }
  \Output{$I_{ava}$}
\end{algorithm}

\section{Human-involved Experiments}
\label{sec:huamn}
For \textbf{human availability inspection}, the annotators who involved in annotating the RF1M dataset are engaged. They followed the same instructions and data format as during dataset construction (\cref{sec:anno}) to mark whether the generated images were available for advertising use.

\begin{figure*}[ht]    \centering
  \includegraphics[width=0.95\columnwidth]{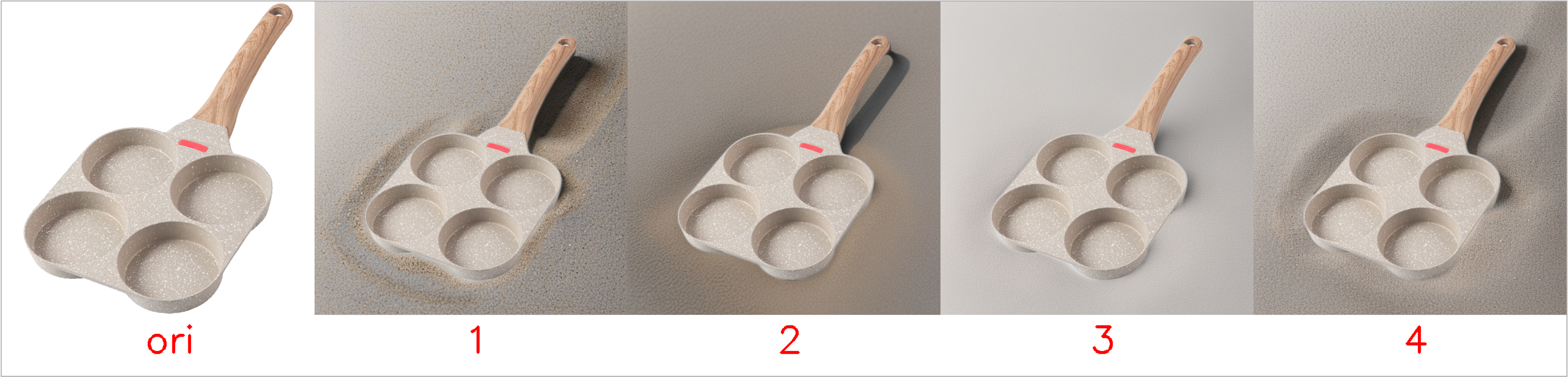}
  \caption{Sample shown to advertising practitioners to assess their preferences.}
  \label{fig:assess}
\end{figure*}

\noindent For \textbf{human preference assessment}, we generated advertising images using different approaches with the same seed. We combined them with the product image in a random sequence to prevent visual fatigue, as shown in \cref{fig:assess}. Each advertising practitioner ranked them based on personal preference, providing a final ranking like ``1423''.

\section{Responsibility to Human Subjects}
\label{sec:respon}
We hired annotators and practitioners to obtain annotations and feedback. We have conducted a thorough review to ensure that the dataset did not include personally identifiable information or offensive content.

\section{Detailed Configuration}
\label{sec:config}
We detail our configuration for image generation in Table \ref{tab:config}.

\begin{table}[h]    \centering
   \addtolength{\tabcolsep}{3.0pt}
   \caption{Image generation configuration. ``RF Interval'' denotes the steps interval to fine-tune the diffusion model.}
   \resizebox{0.85\textwidth}{!}{
       \begin{tabular}{cc|cc}
           \Xhline{1px}
           Image Size & $512\times512$ & Clip Skip & 2 \\
           \hline
           Sampler  & DDIM   & Total Steps  & 40 \\
           \hline
           LoRA Scale  & 0.8   & ControlNet Scale  & 1.0 \\
            \hline

           Guidance Scale  & 7.5   & RF Interval  & 30-40 \\
           \hline
 \multicolumn{4}{c}{\makecell[c]{\textbf{Prompt Example}: ``An item sitting on a wooden table, \\(outdoor background: snowy mountain environment, heavy snow, snowflakes), \\depth of field, close-up, best quality, rich detail, 8k.''}} \\
          \hline
           \multicolumn{4}{c}{\makecell[c]{\textbf{Negative Prompt}: ``text, username, logo, (low quality, worst quality:1.4),\\ (bad anatomy), (inaccurate limb:1.2), bad composition, inaccurate eyes, \\extra digit, fewer digits, (extra arms:1.2), watermark, multiple moles,\\ mole on body, drawing, painting, crayon, sketch, graphite, impressions.''}} \\
           \Xhline{1px}

       \end{tabular}    }    
       
       \label{tab:config}
\end{table}

\section{More Visualization Results}
\label{sec:visual}
\subsection{Image Collapse}

\begin{figure*}[ht]    \centering
  \includegraphics[width=0.95\columnwidth]{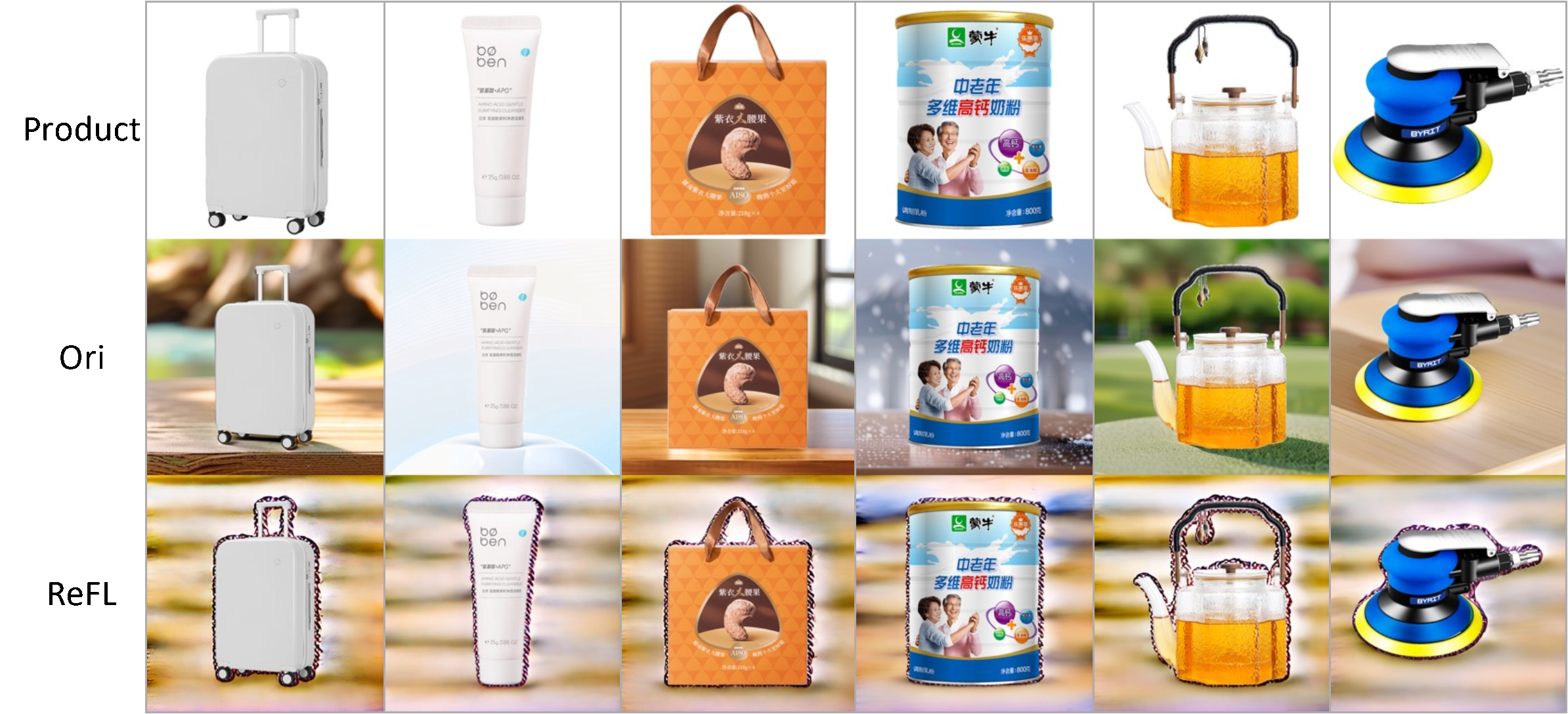}
  \caption{Collapsed images generated by ReFL.}
  \label{fig:hack}
\end{figure*}

In \cref{fig:hack}, we display collapsed images generated by ReFL. Despite varying product images and prompts, the outputs are uniformly degraded and filled with strange textures, similar to adversarial samples \cite{Goodfellow2015Adver, Kurakin2017ADVER}. Thus the direct gradient backpropagation without regularization can be interpreted as training the diffusion model to produce images that deceive the RFNet.

\subsection{Generalization Results}
\begin{figure*}[t]    \centering
  \includegraphics[width=0.95\columnwidth]{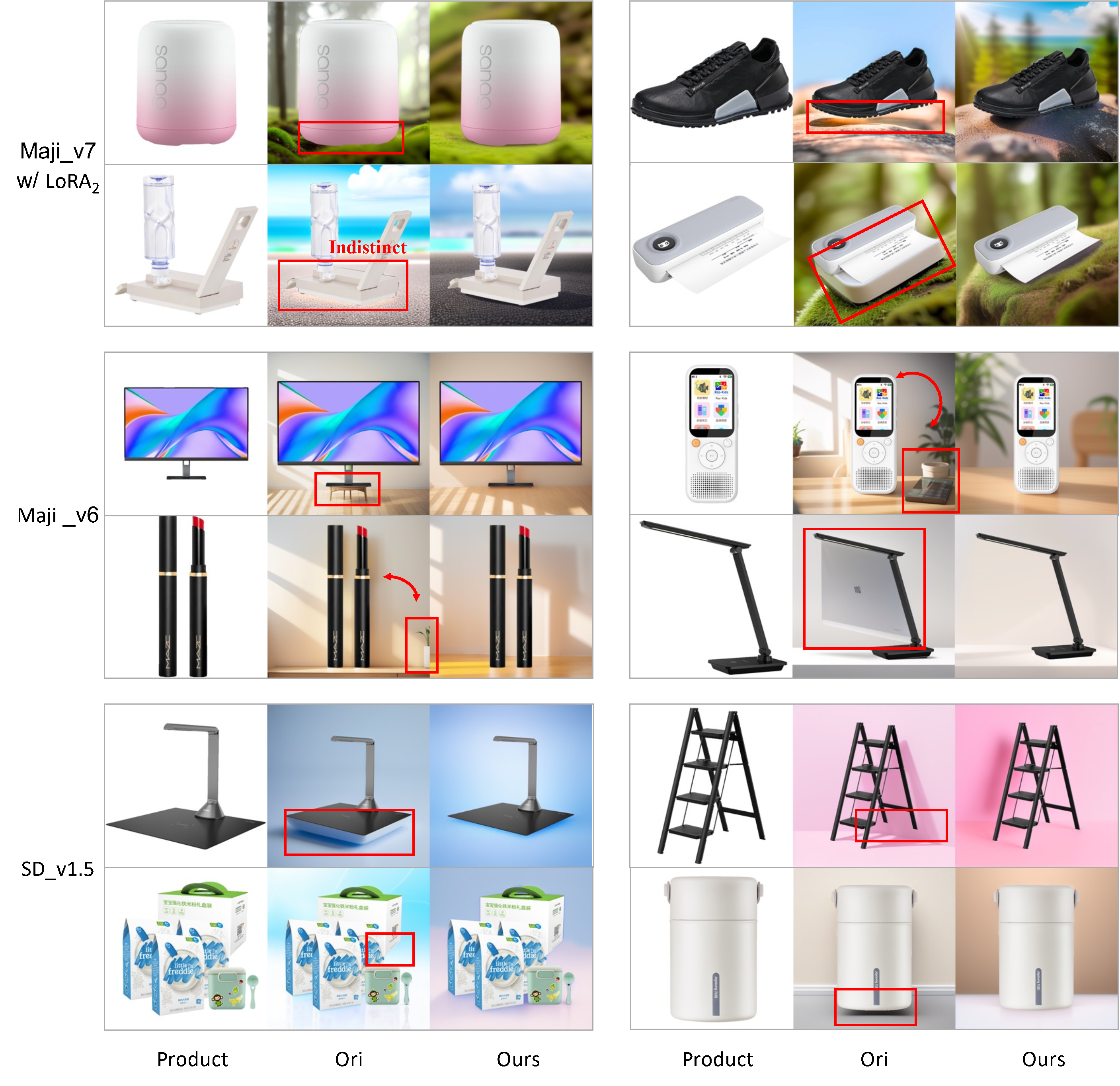}
  \caption{Comparison of the original model and our fine-tuned model using different LoRA and diffusion models.}
  \label{fig:genera}
\end{figure*}

Fig. \ref{fig:genera} shows the consistency of background impressions while the available rate improves with different LoRA/diffusion models. 

\subsection{Integration with Other Feedback}

\begin{figure*}[ht]    \centering
  \includegraphics[width=0.9\columnwidth]{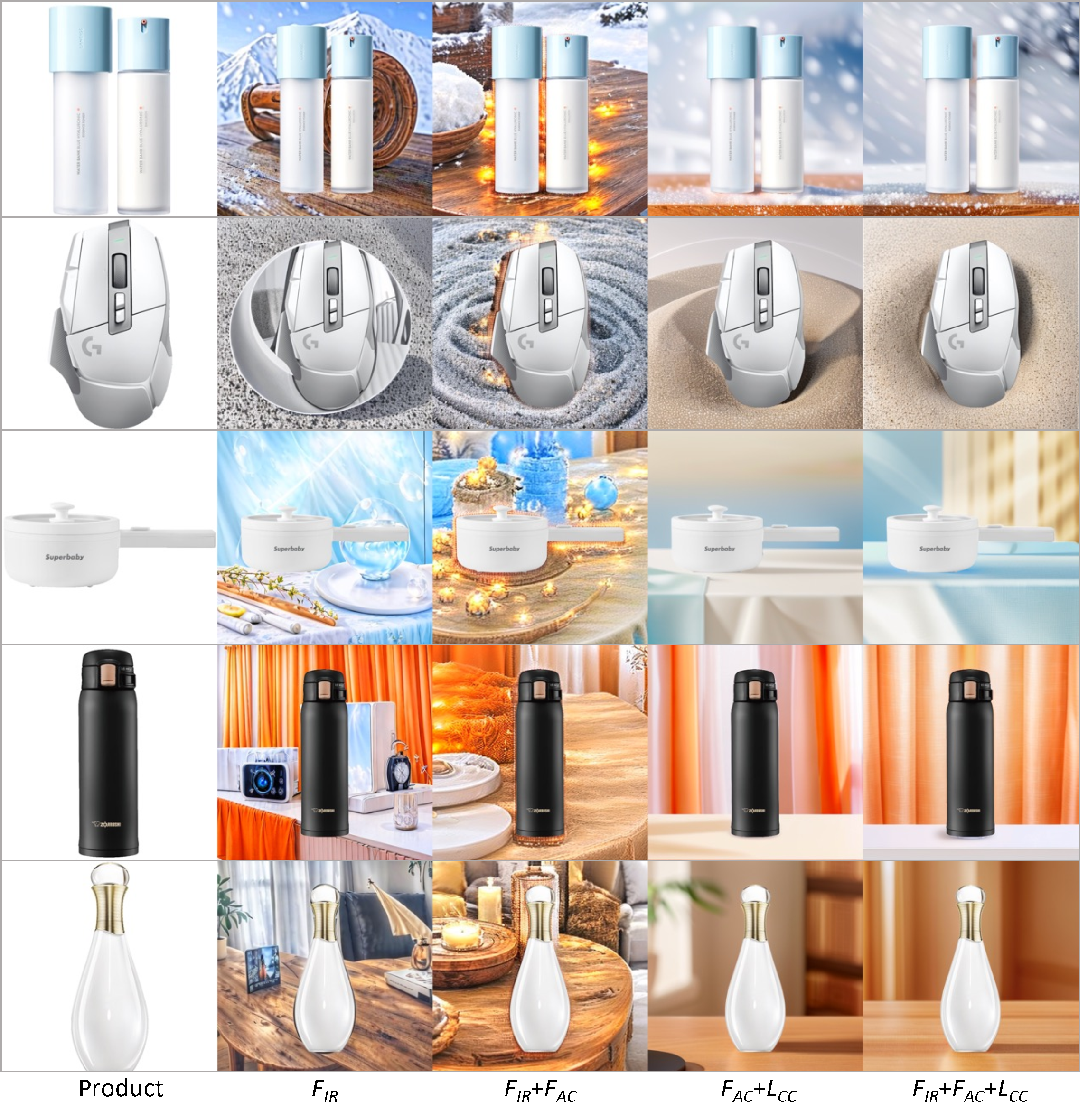}
  \caption{Comparison of different feedback combination strategies.}
  \label{fig:withir}
\end{figure*}

We present further examples with different feedback combination strategies in \cref{fig:withir}. Integrating $F_{IR}$ without $L_{CC}$ leads to overly detailed backgrounds. Combining $F_{IR}$ and $L_{CC}$ preserves background aesthetics, highlighting the role of $L_{CC}$ to maintain the text prompt conditions in backgrounds.

\subsection{Comparison with Other Refining Approaches}
\label{sec:compar}
\begin{figure*}[ht]    \centering
  \includegraphics[width=1.0\columnwidth]{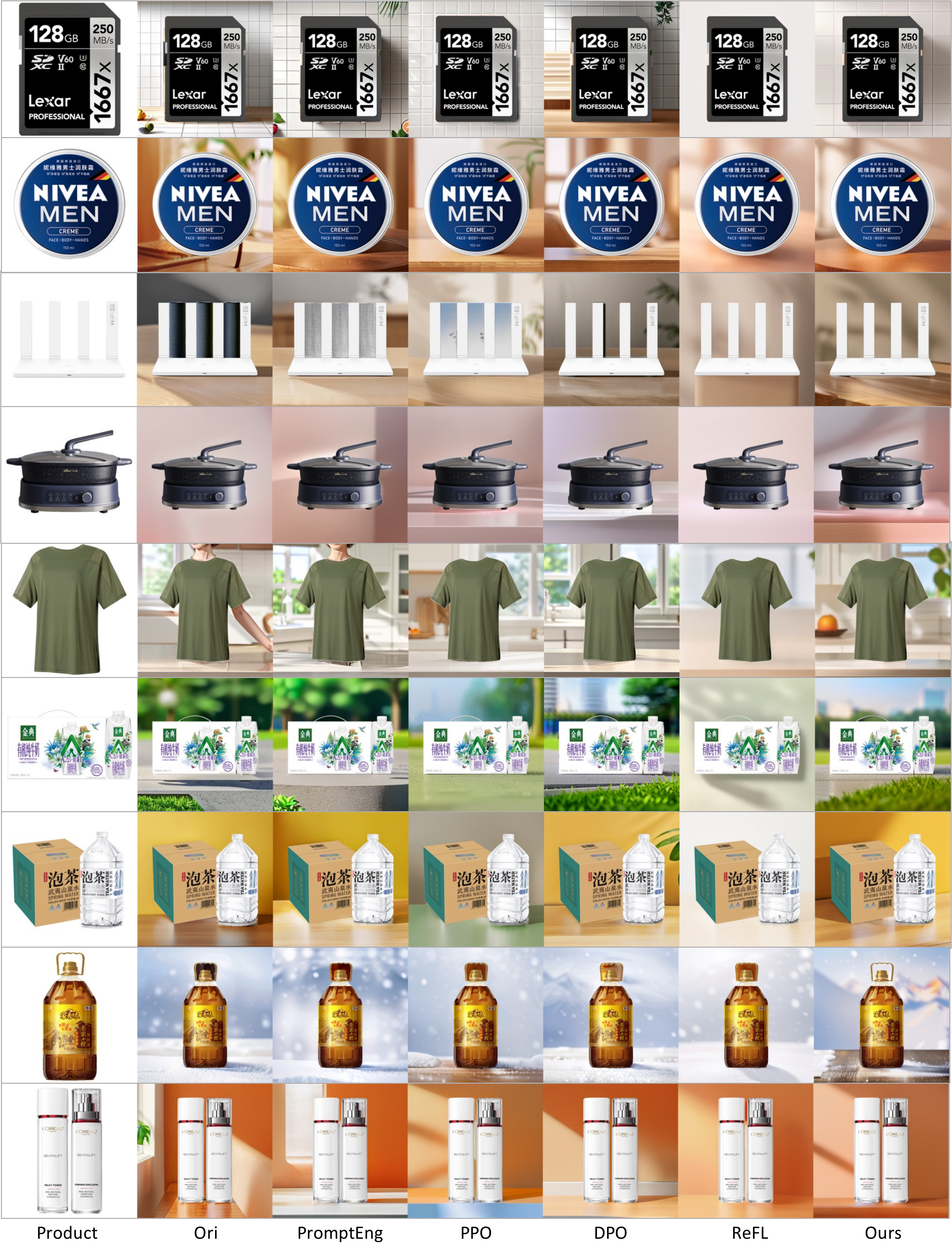}
  \caption{Generated advertising images using different approaches.}
  \label{fig:multi1}
\end{figure*}

\begin{figure*}[ht]    \centering
  \includegraphics[width=1.0\columnwidth]{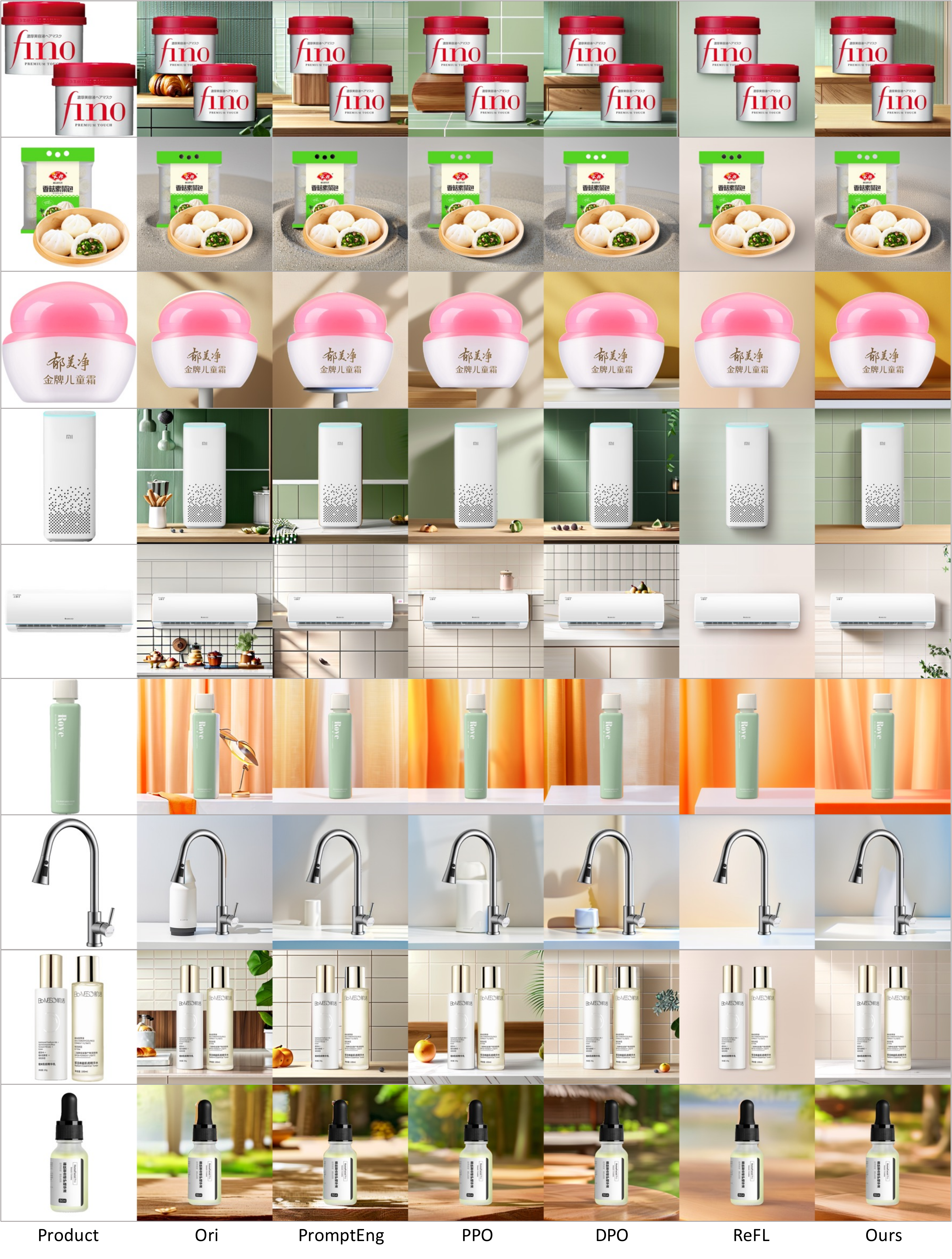}
  \caption{Generated advertising images using different approaches.}
  \label{fig:multi2}
\end{figure*}

Fig. \ref{fig:multi1} and Fig. \ref{fig:multi2} showcase advertising images generated by different approaches. Our approach yields images with a high available rate for advertising purposeS while maintaining appealing visuals.

\section{Ethical Concerns}
\label{sec:ethi}
Regarding image manipulation and automated advertising image generation, there's a risk of producing content that could be unethical or illegal, such as violating individual portrait rights or creating discriminatory content. Thus, these technologies require stricter oversight. In our scenario, we carefully select all product images to ensure neutrality and compliance with advertising standards. The prompts for background generation are also meticulously crafted to produce content that is neutral and meets regulatory requirements.

As for our RF1M dataset, all product information is sourced from an e-commerce platform and has undergone pre-review by both the merchants and the platform. The prompts for generating advertising images are created by professional advertising practitioners. During the generation process, we use Stable Diffusion's official safety checker to filter out NSFW content. Finally, professionals review and screen the dataset, ensuring it is free from bias or offensive content and complies with relevant local laws.

In terms of future applications, while these technologies offer significant benefits in efficiency and personalization, they also pose risks related to misinformation. It's vital to maintain high transparency levels to address these concerns. This includes clearly labeling AI-generated images, allowing consumers to differentiate between AI-generated and human-created content. Furthermore, developing and adhering to business standards and ethical guidelines is crucial to ensure AI's use in advertising upholds truthfulness and honesty. The automation of creative tasks could also alter the job market in creative industries. Rather than seeing AI as a replacement for human creativity, we envision a collaborative approach where AI serves as a tool for creative professionals, helping the workforce stay competitive and ensuring human creativity remains vital in advertising.

\section{Future Work}
\label{sec:future}
Our research currently emphasizes the availability of generated images. Moving forward, we plan to incorporate additional types of feedback to refine diffusion models. For example, click-through rate (CTR) of different images reveals the customers' preferences. Our future work will explore utilizing the CTR to enhance the appeal of generated advertising images.

Additionally, we have found that conventional metrics for evaluating generated images are infeasible for our specific use case. Given that the central focus of advertising image is the identical product, traditional metrics like FID do not provide an accurate assessment of image quality, \ie, different approaches yield similar FIDs despite varying levels of visual attractiveness. For example, ReFL scores an FID of 18.52 compared to our approach's 16.26 in \cref{sec:compar}. Therefore, we intend to develop a more appropriate evaluation metric for assessing the quality of generated backgrounds.

\end{document}